\documentclass[lettersize,journal]{IEEEtran}
\usepackage[misc]{ifsym}
\usepackage{amsmath,amsfonts}
\usepackage{algorithmic}
\usepackage{algorithm}
\usepackage{array}
\usepackage[caption=false,font=normalsize,labelfont=sf,textfont=sf]{subfig}
\usepackage{textcomp}
\usepackage{stfloats}
\usepackage{url}
\usepackage{verbatim}
\usepackage{graphicx}
\usepackage{cite}
\usepackage{tikz}
\usepackage{comment}
\usepackage{amssymb} 
\usepackage{bm}
\usepackage{multirow}
\usepackage{mathtools}
\usepackage{color}
\usepackage{booktabs}
\usepackage{enumitem}

\newcommand{\best}[1]{\textbf{#1}}
\newcommand{\secondbest}[1]{\underline{#1}}

\renewcommand\arraystretch{1.11}  

\begin{document}

\title{CoDA: Color Distribution Probing for Efficient and Generalizable AI-Generated Image Detection}

\author{Zexi Jia,~\IEEEmembership{Member,~IEEE}, 
        Zhiqiang Yuan, 
        Xiaoyue Duan,
        Jinchao Zhang$^*$, 
        Jie Zhou, 
        and Anil K. Jain,~\IEEEmembership{Fellow,~IEEE}%
\thanks{Corresponding author: Jinchao Zhang (E-mail: dayerzhang@tencent.com).}%
\thanks{Zexi Jia, Zhiqiang Yuan, Xiaoyue Duan, Jinchao Zhang, and Jie Zhou are with Tencent WeChat AI, Beijing, China. 
E-mail: \{zexijia, seraphyuan, lilmoonduan, dayerzhang, withtomzhou\}@tencent.com.}%
\thanks{Anil K. Jain is with the Department of Computer Science and Engineering, 
Michigan State University, East Lansing, MI 48824, USA. 
E-mail: jain@egr.msu.edu.}}

\maketitle

\begin{abstract}
AI-generated image detection faces a persistent trade-off between generalization and efficiency: lightweight artifact-based methods often degrade on unseen generators or domains, whereas more robust large-scale models are computationally expensive. Meanwhile, existing benchmarks mainly focus on cross-model evaluation in photorealistic settings, leaving cross-domain robustness underexplored. To address this gap, we introduce FakeForm, a large-scale benchmark with approximately 370,000 images across 62 diverse domains for both cross-model and cross-domain evaluation. Motivated by this broader setting, we revisit color-distribution probing as an efficient complementary cue for AI-generated image detection. We observe that, especially for photographic content, real photographs tend to exhibit smoother and more stable color patterns, whereas synthetic images often show characteristic color imbalances introduced by neural generation. Based on this observation, we propose CoDA, a compact 1.48M-parameter detector built on a Noise-Quantization Probe, together with a theoretical analysis linking probe responses to color non-uniformity. Experiments show that CoDA achieves state-of-the-art performance on standard benchmarks and the best results on the challenging cross-domain evaluation of FakeForm, while remaining highly competitive in cross-model photorealistic settings. These results suggest that persistent generative artifacts can provide a practical foundation for efficient and robust AI-generated image detection. The models and FakeForm benchmark will be made publicly available.
\end{abstract}


\begin{figure*}[!t]
\begin{center}
\includegraphics[width=0.8\linewidth]{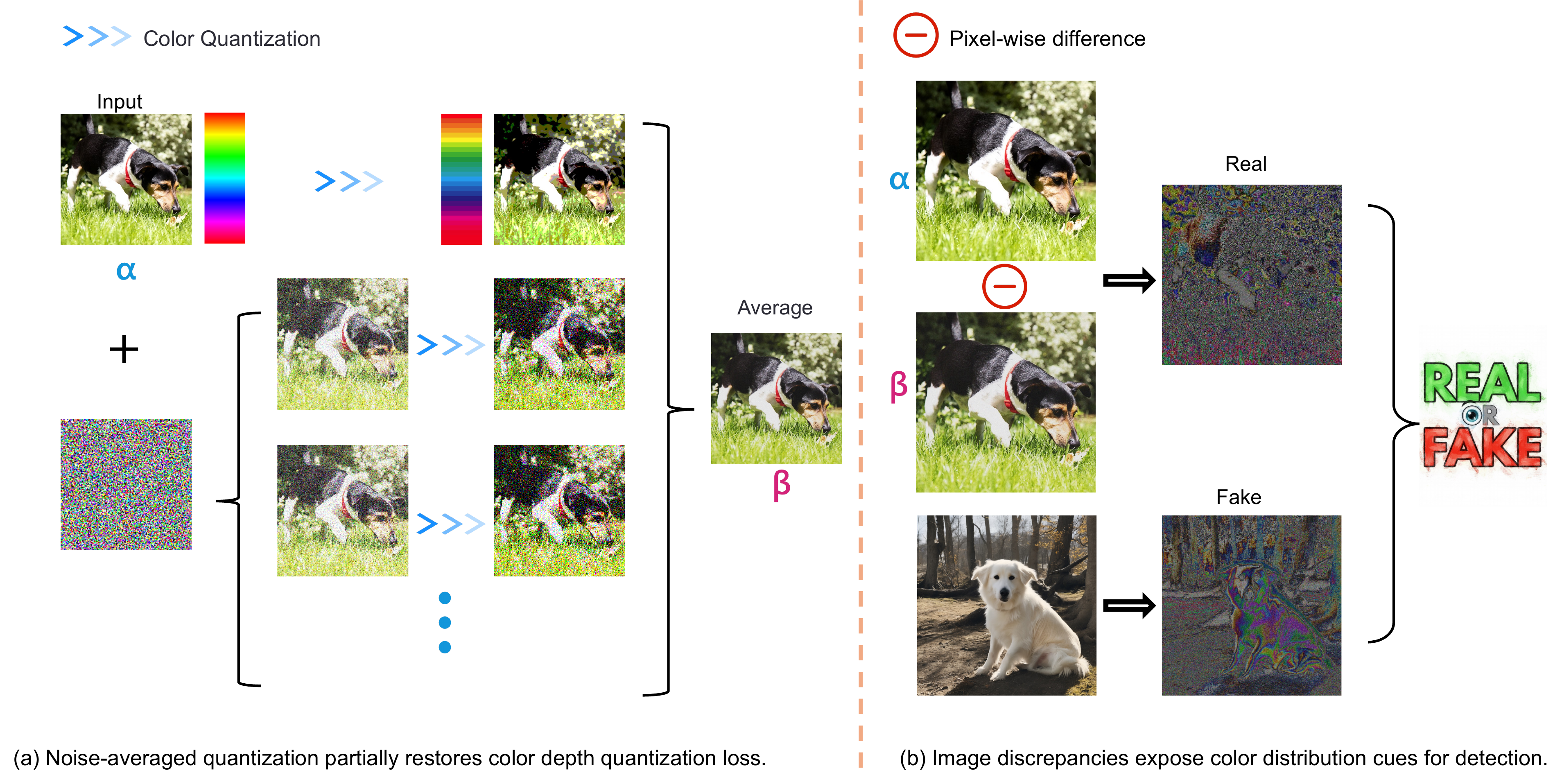}
\end{center}
   \caption{The CoDA detection method exploits systematic differences in color distribution between real and AI-generated images. (a) Our Noise-Quantization Probe measures color distribution stability by adding Gaussian noise to the input, applying color quantization, and averaging multiple samples to partially restore color depth. (b) Real photographs tend to exhibit smoother color distributions and smaller probe-induced changes, while AI-generated images often show more visible color shifts due to non-uniform distributions. The difference maps (magnified 20$\times$ for visualization) provide a useful signal for detection.}
\label{fig:intro}
\end{figure*}

\section{Introduction}
The rapid progress of generative models, including generative adversarial networks (GANs)~\cite{karras2020analyzing}, diffusion models~\cite{ho2020denoising,podell2023sdxl}, autoregressive models~\cite{ramesh2021zero,yu2022scaling}, and related paradigms, has enabled highly realistic synthetic media. While these technologies support many creative applications, they also create serious risks for misinformation, fraud, and the erosion of public trust. Robust AI-generated image detection has therefore become an important practical problem.

Existing detectors face a fundamental trade-off. Low-level forensic methods, including image-based~\cite{wang2020cnn,nataraj2019detecting,guarnera2020deepfake,yu2022responsible} and frequency-based~\cite{frank2020leveraging,tan2024frequency,verdoliva2020media} approaches, are efficient but often rely on model-specific artifacts and degrade on unseen generators or visual domains. Recent large vision-language-model (VLM)-based methods~\cite{ojha2023towards,liu2024forgery,yan2024sanity,zhou2025aigi,jia2025visual} generally offer stronger generalization, but their computational cost is high and their performance can still drop on domains underrepresented in training, such as medical or technical imagery. As a result, current methods still struggle to jointly achieve robustness, efficiency, and broad generalization.

A second limitation lies in current evaluation practice. Most existing benchmarks emphasize cross-model generalization within the relatively narrow regime of photorealistic imagery. In practice, however, detectors must operate across a much broader range of domains, including artistic media, technical diagrams, scientific imagery, and other non-photographic content. Strong performance on conventional benchmarks therefore does not necessarily imply strong cross-domain robustness.

In this paper, we address this problem from two complementary directions. First, we introduce \emph{FakeForm}, a large-scale benchmark designed to evaluate both model and domain generalization. Second, we investigate a lightweight detection paradigm based on color-distribution probing. Human observers often describe synthetic images as ``overly vibrant'' or as having an ``artificial cinematic quality.'' Motivated by this observation, we analyze color statistics and find a recurring color-distribution artifact: AI-generated images tend to exhibit more non-uniform color distributions than real photographs~\cite{fang2026too,jia2026styledecoupler}. This difference can be partly explained by their distinct formation processes. Real photographs are shaped by camera Image Signal Processing (ISP) pipelines, including white balance~\cite{finlayson2004intrinsic}, color correction~\cite{vrhel1992color}, and gamma correction~\cite{foley1982fundamentals}, whereas generative models optimize for semantic fidelity without such physical constraints. This explanation is most direct for photographic content; for other domains, the same cue should be interpreted more cautiously as a broader empirical signal.

To exploit this color-distribution characteristic, we introduce the Noise-Quantization Probe, which reveals distribution differences through controlled stochastic perturbation. As shown in Figure~\ref{fig:intro}, the probe repeatedly adds Gaussian noise, applies color quantization, and averages the resulting samples. Real photographs tend to be more stable under this process, while AI-generated images exhibit larger reconstruction errors and more structured residuals. We provide a theoretical analysis linking this reconstruction error to color-distribution non-uniformity. Based on this observation, we propose CoDA, a lightweight detector that combines the probe signal with spatial image features in a compact 1.48M-parameter network. Rather than relying on color alone, CoDA uses the color branch as a complementary cue jointly with the image branch.

To evaluate these ideas rigorously, we introduce FakeForm, a benchmark designed to expose both model and domain generalization. The model axis covers a diverse set of generators spanning diffusion, flow-matching, autoregressive, and unified multimodal paradigms. The domain axis contains approximately 370,000 images across 62 diverse domains, including categories that are largely absent from prior AIGC detection benchmarks, such as CT scans, depth maps, and flowcharts. FakeForm is further enriched with a large-scale human study comprising more than 760,000 judgments with realism scores and textual justifications, enabling analysis of both machine and human perceptual difficulty.

Experiments on FakeForm reveal a substantial gap between in-domain and cross-domain performance. State-of-the-art detectors often exceed 95\% accuracy on in-domain photorealistic data but fall below 75\% on more challenging cross-domain content. In contrast, CoDA achieves 91.0\% average accuracy on cross-model evaluation over photorealistic content and 77.7\% average accuracy on the more challenging cross-domain setting, while requiring orders of magnitude fewer computational resources than VLM-based baselines. These results suggest that persistent generative artifacts can provide a practical basis for detectors that balance robustness and efficiency.

Our contributions are summarized as follows:
\begin{itemize}
    \item We introduce FakeForm, a comprehensive benchmark for AI-generated image detection with approximately 370,000 images across 62 diverse domains, together with large-scale human perceptual annotations for rigorous cross-domain and cross-model evaluation.
    \item We identify and analyze a color-distribution cue in AI-generated images, and show that it provides a useful complementary signal for detection.
    \item We propose CoDA, a lightweight detection method based on a principled Noise-Quantization Probe that transforms color-distribution irregularities into robust signals.
    \item We demonstrate on standard benchmarks and FakeForm that CoDA achieves strong generalization while maintaining high efficiency.
\end{itemize}

\section{Related Work}

\subsection{Synthetic Image Detection}
Synthetic image detection methods can be broadly divided into two lines of work. The first focuses on \textbf{low-level forensic artifacts} introduced by the generation process. Early studies examined model-specific fingerprints in the pixel domain~\cite{yu2019attributing}, and general-purpose CNNs such as Xception were also evaluated for this task~\cite{rossler2019faceforensics++}. Later work extended this idea to the frequency domain by exploiting spectral artifacts introduced by generator architectures~\cite{qian2020thinking,jeong2022bihpf}. However, the rise of diffusion models has weakened the effectiveness of many such signatures~\cite{ricker2022towards,jia2026evaluating}. More recent methods within this line use generative models themselves as detectors; for example, DIRE~\cite{wang2023dire} leverages diffusion-model reconstruction error, although its transfer to GAN-generated images remains limited. In general, these methods are efficient and interpretable, but their reliance on transient architectural artifacts can limit robustness.

The second line focuses on \textbf{high-level semantic and distributional anomalies} using large-scale pre-trained models. CLIP-based representations have shown that synthetic images can remain distinguishable even when obvious low-level artifacts are weak~\cite{radford2021learning}. Ojha et al.~\cite{ojha2023towards} demonstrated that a simple classifier on CLIP features can generalize surprisingly well. Subsequent work has strengthened this paradigm through more specialized adaptations, such as FatFormer~\cite{liu2024forgery}, Forensics Adapter~\cite{cui2024forensicsadapter}, and InfoFD~\cite{qin2025infofd}. Recent large-model detectors further improve generalization, but they also substantially increase computational cost, which limits practical deployment at scale. These methods are generally more robust than low-level forensic approaches, but they are also less transparent and much more expensive.

Taken together, these two lines of work reflect a persistent trade-off: forensic methods are efficient but often brittle under distribution shift, whereas semantic methods are more robust but substantially more costly. Our work is motivated by the possibility that color-distribution probing can occupy a useful middle ground. Rather than replacing semantic reasoning, we study whether it can provide an efficient \emph{complementary} cue that transfers more effectively across generators and domains.

\subsection{Systematic Color Distortion in Generative Models}

Color irregularities have been repeatedly observed in modern generative models, although their forms and causes vary across architectures. This makes color a meaningful direction for synthetic image detection, especially when one seeks cues that are lightweight and complementary to image-level semantics. Although many studies aim to improve photorealism, color distortions persist as consequences of model architecture and optimization.

In GANs, the trade-off between fidelity and diversity often leads to oversaturation, a tendency amplified by techniques such as the \textit{truncation trick}~\cite{brock2018large}. Architectural choices, including the AdaIN layers in early StyleGAN models, have also been shown to disrupt color statistics~\cite{karras2020analyzing}. In diffusion models, Classifier-Free Guidance (CFG)~\cite{ho2022classifier} is a major source of oversaturation at high guidance strengths, motivating a line of work on mitigation strategies such as dynamic thresholding~\cite{saharia2022imagen} and more recent norm-preserving guidance methods~\cite{jin2025angle,sadat2024eliminating,lin2024wacv}. Autoregressive image generators exhibit related distortions through different mechanisms. Their sequential generation is affected by quantization in discrete visual tokenizers~\cite{van2017neural} and by exposure bias~\cite{ranzato2015sequence}, so early prediction errors can propagate into large-scale hue shifts or color casts. Recent autoregressive generators such as VAR~\cite{tian2024var}, Infinity~\cite{han2025infinity}, HART~\cite{tang2025hart}, and Hi-MAR~\cite{zheng2025himar} further highlight the need to evaluate detection beyond the traditional GAN--diffusion setting. In parallel, D3QE~\cite{zhang2025d3qe} studies detection specifically in the autoregressive regime, reflecting growing interest in this problem.

Overall, existing studies suggest that color distortion is a recurring artifact across major generation paradigms, but its exact form is architecture-dependent rather than universal. This is consistent with the perspective adopted in this paper: for photographic content, color-distribution differences admit a more direct ISP-related interpretation, whereas for non-photographic domains they are better understood as a broader empirical cue. This motivates our use of color-distribution probing as a lightweight and complementary signal for AI-generated image detection.

\section{Analysis of Generative Color Distortion}
\label{sec:framework}

\begin{figure*}[!t]
\begin{center}
\includegraphics[width=0.85\linewidth]{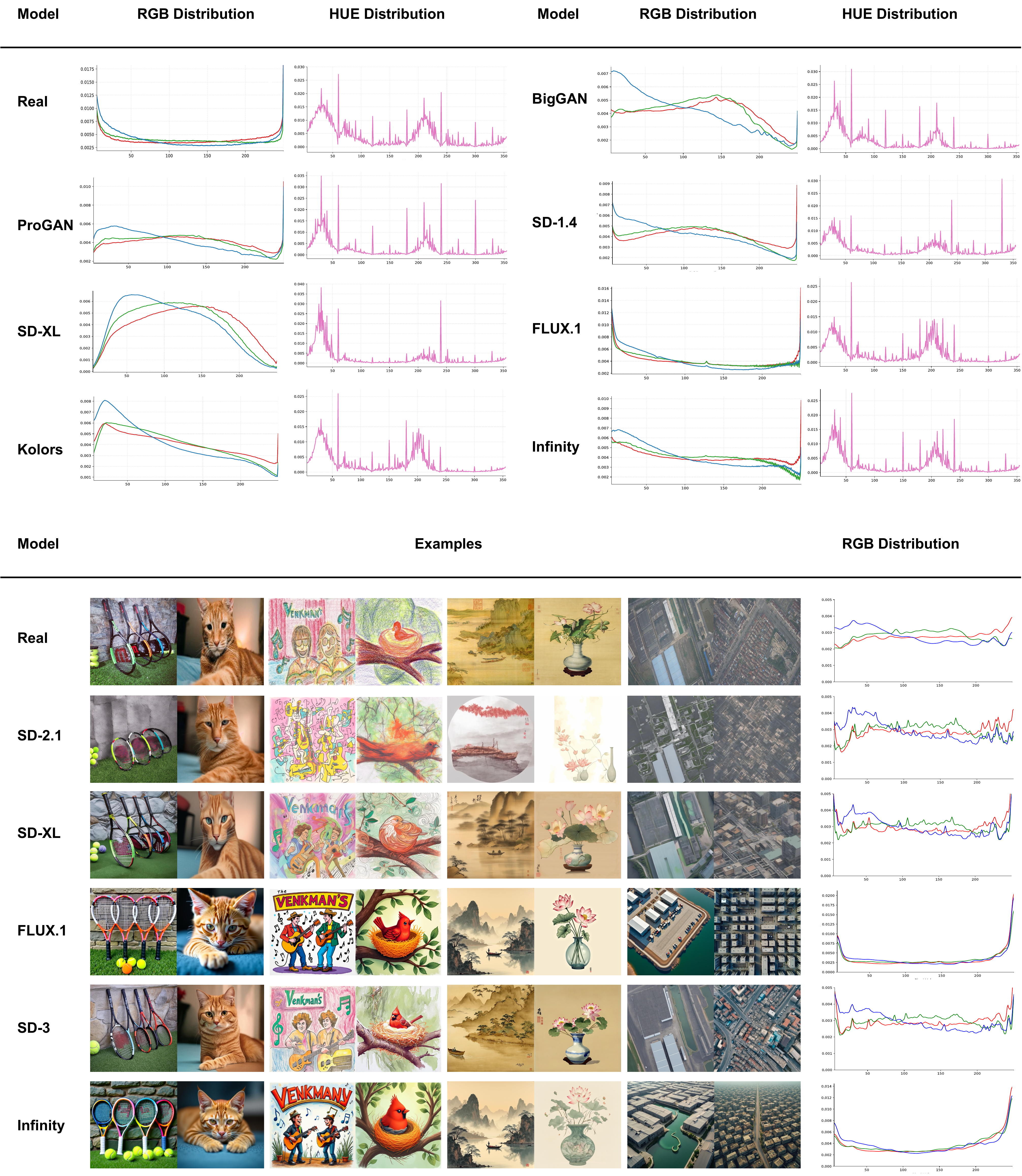}
\end{center}
\caption{
Color statistics of real and synthetic images.
The top panels show aggregated RGB and hue distributions from large-scale real and synthetic image sets: real photographs tend to exhibit smooth and broad distributions, whereas generative models produce sharper peaks concentrated in narrower regions.
The bottom panels show representative one-to-one real--fake pairs generated by an image-to-image ControlNet pipeline.
Together, they show that the color imbalance reflects a systematic perceptual gap between real and synthetic content rather than merely scene variation.
}
\label{fig:count}
\end{figure*}

\subsection{Empirical Evidence of Synthetic Color Inhomogeneity}

Color distortion, including oversaturation, hue shifts, and banding artifacts, is common in modern generative models. We therefore begin by asking a simple question: do synthetic images exhibit systematically different color statistics from real photographs? We observe a consistent discrepancy, and the effect remains visible both in large-scale aggregate statistics and in controlled one-to-one comparisons.

Human observers often describe synthetic images as ``unreal'' or ``overly sanitized'' even when obvious structural artifacts are absent. A key factor is the statistical inhomogeneity of their color distributions. Compared with real photographs, synthetic images frequently exhibit unnaturally concentrated or muted palettes and lack the subtle interplay of hues and tones characteristic of natural scenes.

To quantify this effect, we conduct a large-scale analysis of color distributions. For real photographs, we randomly sample 100{,}000 examples from ten public datasets, including COCO~\cite{lin2014coco}, ImageNet~\cite{Russakovsky2015ImageNet}, and LSUN~\cite{yu2015lsun}. For synthetic images, we generate over 60{,}000 samples per model using a curated subset of prompts from HPDv2~\cite{wu2023human}. Our study covers GANs (BigGAN, ProGAN), diffusion models (SD-1.4, SD-XL), flow-based models (FLUX), and others (Kolors, Infinity). For each image, we compute normalized histograms of RGB intensities and hue values (HSV), and summarize them with histogram entropy and a smoothness measure based on the $\ell_2$ norm of adjacent-bin differences.

The top panels of Figure~\ref{fig:count} summarize the aggregated statistics computed from these large-scale real and synthetic image sets. Real photographs exhibit smooth and broad RGB distributions together with relatively uniform hue distributions, whereas generated images show more irregular patterns. Their RGB histograms typically contain sharper peaks, lower entropy, and larger smoothness penalties, indicating that pixel values concentrate in narrower ranges. Hue distributions likewise exhibit isolated spikes corresponding to a few dominant hues. These trends are consistently observed across GANs, diffusion models, and flow-based generators.

To further verify that this phenomenon is not merely caused by scene variation, we additionally construct a controlled paired setting, illustrated in the bottom panels of Figure~\ref{fig:count}. Specifically, we select real images from 10 representative FakeForm domains and, for each real image, generate a semantically matched synthetic counterpart using an image-to-image ControlNet pipeline. In this way, each fake sample is generated in a one-to-one manner from its corresponding real image. For each generator, we construct 2{,}000 such paired samples and recompute the corresponding color statistics on these matched sets. Although the real and generated images are aligned in semantic content and overall composition, clear RGB-distribution discrepancies remain evident under this controlled setting.

Taken together, the two parts of Figure~\ref{fig:count} provide complementary evidence. The large-scale aggregate statistics show that color-distribution discrepancies are systematic at the population level, while the one-to-one ControlNet pairs show that the effect persists even after substantially reducing content variation. These findings indicate that current generative models, despite their high perceptual fidelity, still under-represent the statistical diversity of real-image color distributions.

\subsection{An Intuitive Manifold-Based Explanation}
\label{ssec:theory_of_divergence}

To interpret these observations, we adopt a manifold-divergence perspective. The key idea is that generative pipelines optimize tractable proxy objectives, and this optimization can push samples away from the low-dimensional manifold of real images. In this process, chromatic directions are often particularly sensitive, making color distortion a natural and recurring form of deviation.

We interpret the observed color inhomogeneity as one instance of a more general phenomenon. Generative models are trained and guided by proxy objectives that do not perfectly match the true data distribution. Aggressive optimization of these objectives can induce off-manifold deviations, and color-related dimensions are especially susceptible.

Let the ambient image space be $\mathcal{X} \subset \mathbb{R}^D$, where $D$ is the number of pixels. The true data distribution $p_{\text{data}}$ is supported on a low-dimensional nonlinear manifold $\mathcal{M} \subset \mathcal{X}$ with intrinsic dimension $d \ll D$. A generative model with parameters $\theta$ learns an approximation $p_{\theta}$ by minimizing a tractable proxy objective $\mathcal{L}_{\text{proxy}}$. During inference, guidance mechanisms often maximize a proxy score function $s: \mathcal{X} \to \mathbb{R}$ tied to $\mathcal{L}_{\text{proxy}}$.

We quantify the deviation of a generated sample $\hat{\mathbf{x}}$ from the natural-image manifold by the \emph{manifold deviation distance}
\begin{equation}
    d_{\mathcal{M}}(\hat{\mathbf{x}}) = \|\hat{\mathbf{x}} - \Pi_{\mathcal{M}}(\hat{\mathbf{x}})\|_2,
\end{equation}
where $\Pi_{\mathcal{M}}: \mathcal{X} \to \mathcal{M}$ denotes the Euclidean projection onto $\mathcal{M}$. Larger $d_{\mathcal{M}}$ indicates stronger violations of natural image statistics and may appear as color distortion, unnatural texture, or structural inconsistency. Among these, chromatic deviations are often especially salient because human vision is highly sensitive to hue and saturation changes.

Guidance typically proceeds through gradient-based updates
\begin{equation}
    \mathbf{x}_{t+1} = \mathbf{x}_t + w \cdot \nabla_{\mathbf{x}} s(\mathbf{x}_t),
\end{equation}
where $w > 0$ is the guidance strength. The score gradient at $\mathbf{x}_t \in \mathcal{M}$ can be decomposed into tangent and normal components:
\begin{equation}
    \nabla_{\mathbf{x}} s(\mathbf{x}_t) = \nabla_{\mathbf{x}}^{\parallel} s(\mathbf{x}_t) + \nabla_{\mathbf{x}}^{\perp} s(\mathbf{x}_t).
\end{equation}
Under a first-order Taylor approximation of the projection operator and for sufficiently small steps, the change in manifold deviation is dominated by the normal component:
\begin{equation}
    d_{\mathcal{M}}(\mathbf{x}_{t+1}) - d_{\mathcal{M}}(\mathbf{x}_t) \approx w \cdot \|\nabla_{\mathbf{x}}^{\perp} s(\mathbf{x}_t)\|_2.
\end{equation}
Thus, large guidance strengths $w \gg 1$ amplify non-tangential components of the score gradient and push samples away from $\mathcal{M}$. When the normal component has substantial projection onto color-related directions, the resulting off-manifold excursion can appear as the distorted color statistics observed empirically.

This manifold-divergence view offers a unified explanation for why color artifacts recur across different generation paradigms. Although the precise form of the deviation depends on the model class and inference procedure, deviations along color-related directions provide a natural account of the statistical patterns observed in Figure~\ref{fig:count}.

\subsection{Architecture-Specific Mechanisms of Color Distortion}
\label{sec:manifestations}

The principle of manifold divergence applies broadly, but its concrete manifestation depends on the architecture and inference procedure. We highlight how this principle leads to characteristic color distortions in diffusion, GAN, and autoregressive models.

\textbf{Diffusion Models.}
High-fidelity diffusion generation commonly uses Classifier-Free Guidance (CFG)~\cite{ho2022classifier}, which constructs an extrapolated noise estimate from a conditional prediction $\epsilon_\theta(\cdot, c)$ and an unconditional prediction $\epsilon_\theta(\cdot, \varnothing)$:
\begin{equation}
    \tilde{\epsilon}_\theta(\mathbf{x}_t,t,c,w) = \epsilon_\theta(\mathbf{x}_t,t,\varnothing) + w \cdot \big(\epsilon_\theta(\mathbf{x}_t,t,c) - \epsilon_\theta(\mathbf{x}_t,t,\varnothing)\big).
\end{equation}
For $w>1$, this extrapolation moves the sample toward regions of lower model probability. Recent work~\cite{sadat2024eliminating} shows that the guidance vector contains components with distinct roles: one largely amplifies existing content, while another introduces new details. At high guidance strength, the amplification component can dominate, often appearing as oversaturation and exaggerated contrast.

In the manifold view, CFG increases the magnitude of $\nabla^{\perp} s$ whenever guidance emphasizes features not faithfully captured by the learned data manifold. When these directions correlate with color and contrast, off-manifold updates naturally produce the color distortions observed in our empirical study.

\textbf{Generative Adversarial Networks.}
In GANs, color distortion is closely tied to how the generator exploits the discriminator. Beyond sampling schemes such as the truncation trick~\cite{brock2018large}, practical pipelines often apply inference-time latent refinement that seeks latent codes $\mathbf{z}$ maximizing the discriminator score $D(G(\mathbf{z}))$:
\begin{equation}
    \nabla_{\mathbf{z}} D(G(\mathbf{z})) = J_G(\mathbf{z})^\top \nabla_G D(G(\mathbf{z})),
\end{equation}
where $J_G(\mathbf{z})$ is the generator Jacobian. If the discriminator relies on predictive but non-robust cues such as elevated chroma or stylized grading, then $\nabla_G D$ tends to point toward regions of stronger saturation or contrast.

When $J_G(\mathbf{z})$ has large singular values aligned with color-sensitive directions, small latent updates can produce large color shifts in image space. This nudges samples into regions with exaggerated color statistics outside the typical real-image manifold.

\textbf{Autoregressive Models.}
Autoregressive models factorize the image distribution as $p(\mathbf{x}) = \prod_i p(x_i \mid \mathbf{x}_{<i})$ over discrete tokens from a VQ-VAE~\cite{van2017neural}. Their color artifacts mainly stem from codebook quantization and exposure bias~\cite{ranzato2015sequence}. Exposure bias causes early prediction errors to propagate and accumulate. An error $\bm{\delta}_i$ at step $i$ contributes to the final error $\bm{\delta}_N$ approximately as
\begin{equation}
    \bm{\delta}_N \approx \sum_{i=1}^{N-1} \mathbf{A}_{N \leftarrow i} \bm{\delta}_i,
\end{equation}
where $\mathbf{A}_{N \leftarrow i}$ is an error propagation matrix.

In many practical tokenization schemes, early tokens encode low-frequency structure, including global luminance and chrominance. A small quantization or prediction error in one of these tokens can therefore bias the global color basis of the image. As this error propagates and is possibly amplified, it may lead to large-scale hue drifts or color casts that affect the whole image.

Across these architectures, the specific mechanisms differ, but the overall pattern is consistent: proxy-driven optimization can introduce deviations in color-related directions that are reflected in the statistics measured by our probe. These mechanisms should be understood as architecture-specific tendencies rather than a single exhaustive explanation.

\section{Method}
\label{sec:method}

\subsection{Motivation and Overview}

Our method is motivated by the observation that the stability of an image under controlled noise and quantization reflects its underlying color distribution. As discussed in Section~\ref{sec:framework}, real and generated images exhibit systematic differences in color statistics, and these differences remain visible even under controlled one-to-one comparisons. This makes color-distribution probing a useful complementary signal for AI-generated image detection.

The key challenge is to quantify color distribution uniformity from a single image. Direct histogram analysis is insufficient because it ignores spatial information and cannot distinguish natural variation from artificial patterns. We therefore seek a mechanism that reveals whether an image contains large regions of color non-uniformity.

We introduce the Noise-Quantization Probe, a simple framework that exposes hidden color-distribution characteristics. The central idea is that the interaction between noise, quantization, and color distribution produces measurable bias patterns: broad and balanced distributions tend to yield perturbation errors that cancel after averaging, whereas peaked or irregular distributions produce more systematic residuals.

\subsection{Noise-Quantization Probe}

Given an RGB image $I \in [0,255]^{H \times W \times 3}$, the probe operates in three steps.

\textbf{Step 1: Noise Injection.}
We generate $R$ perturbed versions by adding Gaussian noise:
\begin{equation}
    I'_r = I + N_r,
\end{equation}
where $N_r \sim \mathcal{N}(0,\sigma^2)$ and $\sigma$ controls the perturbation strength.

\textbf{Step 2: Quantization.}
Each perturbed image is quantized as
\begin{equation}
    Y_r = \mathrm{round}(I'_r).
\end{equation}

\textbf{Step 3: Restoration.}
The restored image is obtained by averaging:
\begin{equation}
    \hat{I} = \frac{1}{R} \sum_{r=1}^{R} Y_r.
\end{equation}

The key output is the difference map
\begin{equation}
    \Delta = I - \hat{I},
\end{equation}
which captures the systematic bias between the original and restored images.

\begin{figure}[t]
\begin{center}
\includegraphics[width=\linewidth]{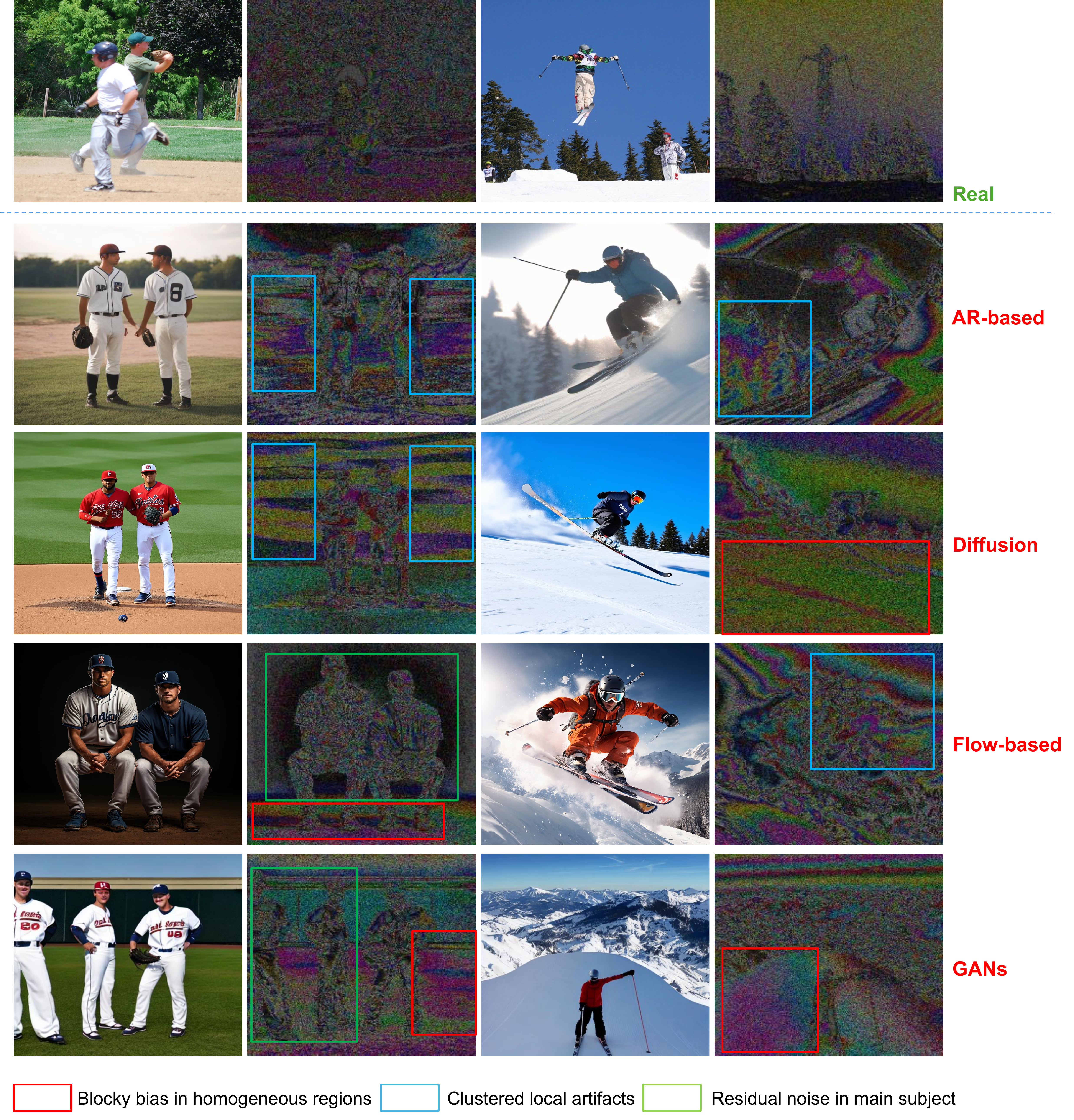}
\end{center}
\caption{Noise-Quantization Probe responses across generative models. Real images (top row) produce relatively uniform difference maps, while synthetic images from different architectures (AR-based, diffusion, flow-based, and GANs) exhibit characteristic artifacts, including blocky bias in smooth regions (red boxes), clustered local artifacts (blue boxes), and residual noise around subjects (green boxes). These patterns provide useful evidence for detecting synthetic images across several generator families.}
\label{fig:show}
\end{figure}

As shown in Figure~\ref{fig:show}, the response of $\Delta$ depends on the underlying color distribution. In images with broad color distributions, perturbation-induced biases tend to cancel during averaging, yielding a small residual. In contrast, when pixel values cluster within restricted ranges, the induced biases become more directional and persist after averaging, producing a larger and more structured difference map.

\subsection{Theoretical Analysis}

We now explain why the difference map is sensitive to color distributions. For clarity, we work on a normalized intensity scale $x \in [0,1)$ obtained by dividing by $L=256$. Consider
\begin{equation}
    Y = \mathrm{round}(x + N), \quad N \sim \mathcal{N}(0,\sigma^2),
\end{equation}
where rounding is to the nearest integer. Away from saturation boundaries, clipping effects are negligible and $Y$ can be treated as taking values in $\mathbb{Z}$.

The conditional probability that $Y=y$ given $x$ is
\begin{equation}
    P(Y = y \mid X = x)
    = \Phi\!\left(\frac{y+0.5 - x}{\sigma}\right)
    - \Phi\!\left(\frac{y-0.5 - x}{\sigma}\right),
\end{equation}
where $\Phi(\cdot)$ is the standard normal CDF. The expected quantized value is
\begin{equation}
    \mathbb{E}[Y \mid X = x] = \sum_{y \in \mathbb{Z}} y \, P(Y = y \mid X = x),
\end{equation}
and the restoration bias is
\begin{equation}
    \mathcal{B}(x;\sigma) = \mathbb{E}[Y \mid X = x] - x.
\end{equation}

\paragraph{Fourier expansion of the bias}
Because quantization is translation-invariant with period $1$, $\mathcal{B}(x;\sigma)$ is a $1$-periodic function of $x$ and admits a Fourier series
\begin{equation}
    \mathcal{B}(x;\sigma) = \sum_{m \in \mathbb{Z}} a_m(\sigma)e^{i2\pi m x},
\end{equation}
with coefficients
\begin{equation}
    a_m(\sigma) = \int_0^1 \mathcal{B}(x;\sigma)e^{-i2\pi m x}\,dx.
\end{equation}
The average bias over one period vanishes, so $a_0(\sigma)=0$. For $m \neq 0$, standard algebra yields
\begin{equation}
    a_m(\sigma) = \frac{i(-1)^m}{m\pi}\exp\!\bigl(-2\pi^2\sigma^2m^2\bigr).
\end{equation}
Since $\mathcal{B}(x;\sigma)$ is real-valued, we have
\begin{equation}
    \mathcal{B}(x;\sigma)
    = \sum_{m=1}^{\infty}\frac{(-1)^m}{m\pi}
      \exp\!\bigl(-2\pi^2\sigma^2m^2\bigr)\sin(2\pi m x).
    \label{eq:bias_fourier}
\end{equation}
The exponential factor suppresses higher harmonics rapidly. For typical noise levels, the first harmonic dominates, yielding
\begin{equation}
    \mathcal{B}(x;\sigma) \approx C(\sigma)\sin(2\pi x), \quad
    C(\sigma)=\frac{1}{\pi}\exp\!\bigl(-2\pi^2\sigma^2\bigr).
    \label{eq:bias_sinusoid}
\end{equation}

\paragraph{Connection to color distributions}
Let $x_{hwc} \in [0,1)$ denote the normalized intensity at position $(h,w)$ and channel $c$, and let $\Delta_{hwc}$ denote the corresponding entry in the difference map. Ignoring residual noise,
\begin{equation}
    \mathbb{E}[\Delta_{hwc} \mid x_{hwc}] \approx \mathcal{B}(x_{hwc};\sigma)
    \approx C(\sigma)\sin(2\pi x_{hwc}).
\end{equation}
Summing over all pixels and channels gives
\begin{equation}
    \sum_{h,w,c}\mathbb{E}[\Delta_{hwc}]
    \approx C(\sigma)\sum_{h,w,c}\sin(2\pi x_{hwc}).
\end{equation}
If color values are uniformly distributed on $[0,1)$, then
\begin{equation}
    \mathbb{E}_{X \sim \mathrm{Unif}[0,1)}[\sin(2\pi X)] = 0,
\end{equation}
so positive and negative contributions cancel and the average difference remains close to zero. For a non-uniform intensity distribution $p_X$, however, the weighted average of $\sin(2\pi X)$ under $p_X$ will in general deviate from zero:
\begin{equation}
    \mathbb{E}_{X \sim p_X}[\sin(2\pi X)] \neq 0 \quad \text{in general},
\end{equation}
which can induce a non-zero bias and hence a larger difference magnitude.

Consequently,
\begin{equation}
    \|\Delta\| \approx \biggl| C(\sigma)\sum_{h,w,c}\sin(2\pi x_{hwc}) \biggr|
\end{equation}
acts as an indicator of color non-uniformity. Natural photographs tend to produce smaller global biases, whereas synthetic images with systematic color bias produce larger and more structured responses. Consistent with Section~\ref{sec:framework}, this analysis is best viewed as principled support for the probe mechanism, especially in photographic settings, rather than as evidence that a single causal explanation governs all domains in FakeForm.

\subsection{Detection Architecture}

\begin{algorithm}[t]
\caption{CoDA inference with the Noise-Quantization Probe}
\label{alg:method}
\begin{algorithmic}[1]
\STATE \textbf{Input:} Image $I \in [0,255]^{H \times W \times 3}$, parameters $\sigma$, $R$
\STATE \textbf{Output:} Detection score $p \in [0,1]$
\STATE
\STATE \textit{// Noise-Quantization Probe}
\STATE Initialize $\hat{I} \leftarrow \mathbf{0}$
\FOR{$r = 1$ to $R$}
  \STATE Sample noise $N_r \sim \mathcal{N}(0, \sigma^2\mathbf{I})$
  \STATE Compute $Y_r \leftarrow \mathrm{clip}(\mathrm{round}(I + N_r),0,255)$
  \STATE Update $\hat{I} \leftarrow \hat{I} + Y_r/R$
\ENDFOR
\STATE Compute difference map $\Delta \leftarrow I - \hat{I}$
\STATE
\STATE \textit{// Feature Extraction and Classification}
\STATE Extract visual features: $\mathbf{f}_v \leftarrow \text{CNN}_v(I)$
\STATE Extract color features: $\mathbf{f}_c \leftarrow \text{CNN}_c(\Delta)$
\STATE Fuse features: $\mathbf{f} \leftarrow [\mathbf{f}_v \,\|\, \mathbf{f}_c]$
\STATE Compute detection score: $p \leftarrow \text{Classifier}(\mathbf{f})$
\STATE \textbf{return} $p$
\end{algorithmic}
\end{algorithm}

\begin{figure}[t]
\begin{center}
\includegraphics[width=\linewidth]{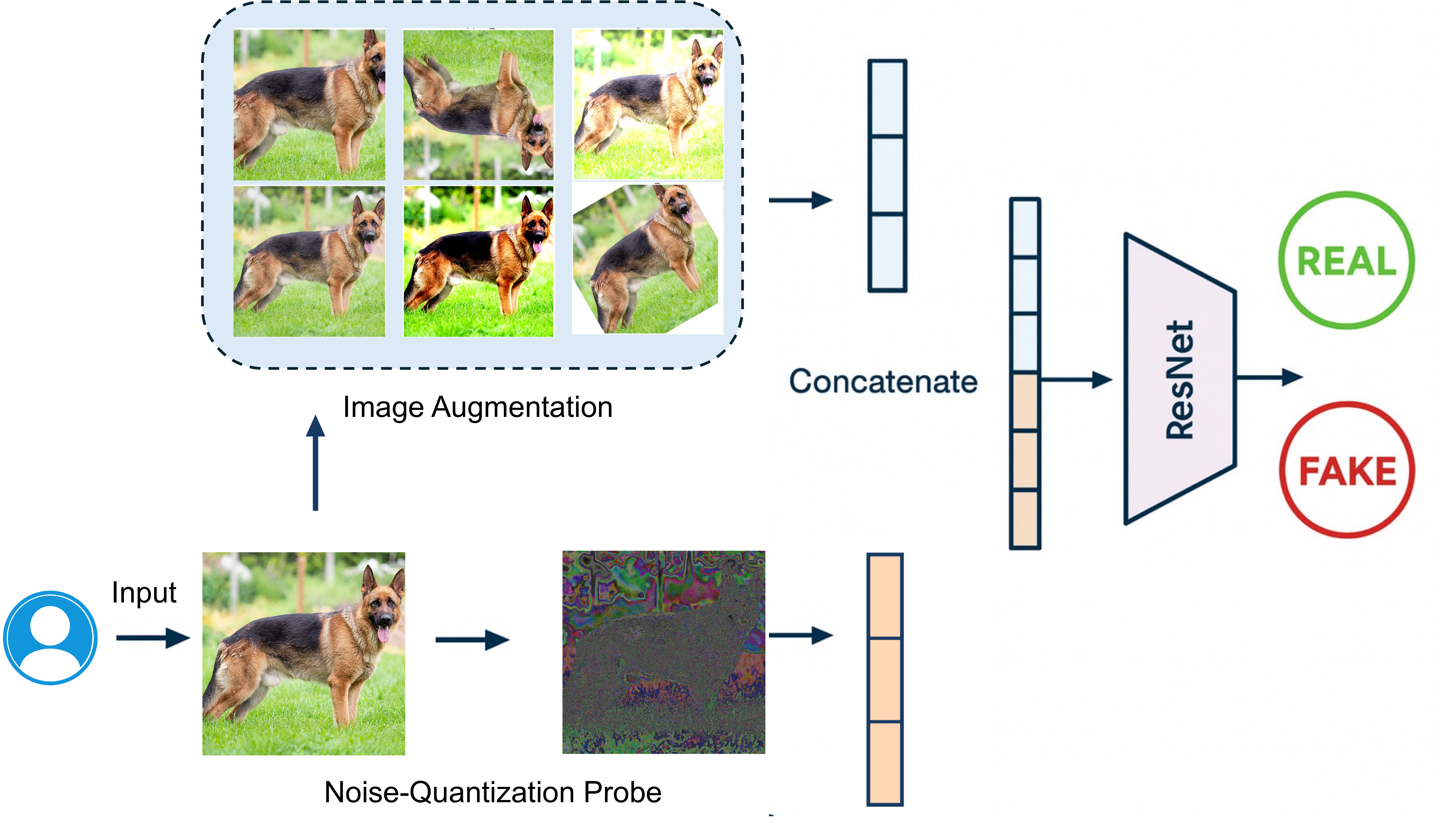}
\end{center}
\caption{Overview of CoDA's architecture. The dual-branch network processes both the original image through the visual branch (top) and the color difference map generated by the Noise-Quantization Probe through the color branch (bottom). During training, comprehensive image augmentations enhance robustness. The concatenated features from both branches are classified by a lightweight classification head to produce the final detection score.}
\label{fig:net}
\end{figure}

The final detector adopts a lightweight dual-branch architecture, illustrated in Figure~\ref{fig:net}. Given an input image $I$, the Noise-Quantization Probe first produces a difference map $\Delta = I-\hat{I}$. The original image is then processed by a visual branch, while the difference map is processed by a color branch. Their features are concatenated and fed to a lightweight classifier to produce the final prediction.

This design combines two complementary sources of evidence. The visual branch captures semantic and textural information, while the color branch extracts probe-induced distributional cues. As a result, CoDA does not rely solely on color; instead, color-distribution probing serves as an additional signal that improves robustness when fused with image features.

To improve robustness, we use standard data augmentation during training, including random resizing, cropping, rotation, horizontal flipping, color jittering, and random rectangular occlusion. These augmentations are applied consistently so that the image branch and color branch remain aligned. Overall, this compact architecture preserves the efficiency advantage of CoDA while enabling effective integration of color-distribution cues into a practical detector.

\section{FakeForm: A Benchmark for Cross-Domain AI-Generated Image Detection}

\begin{figure*}[!t]
\begin{center}
\includegraphics[width=\linewidth]{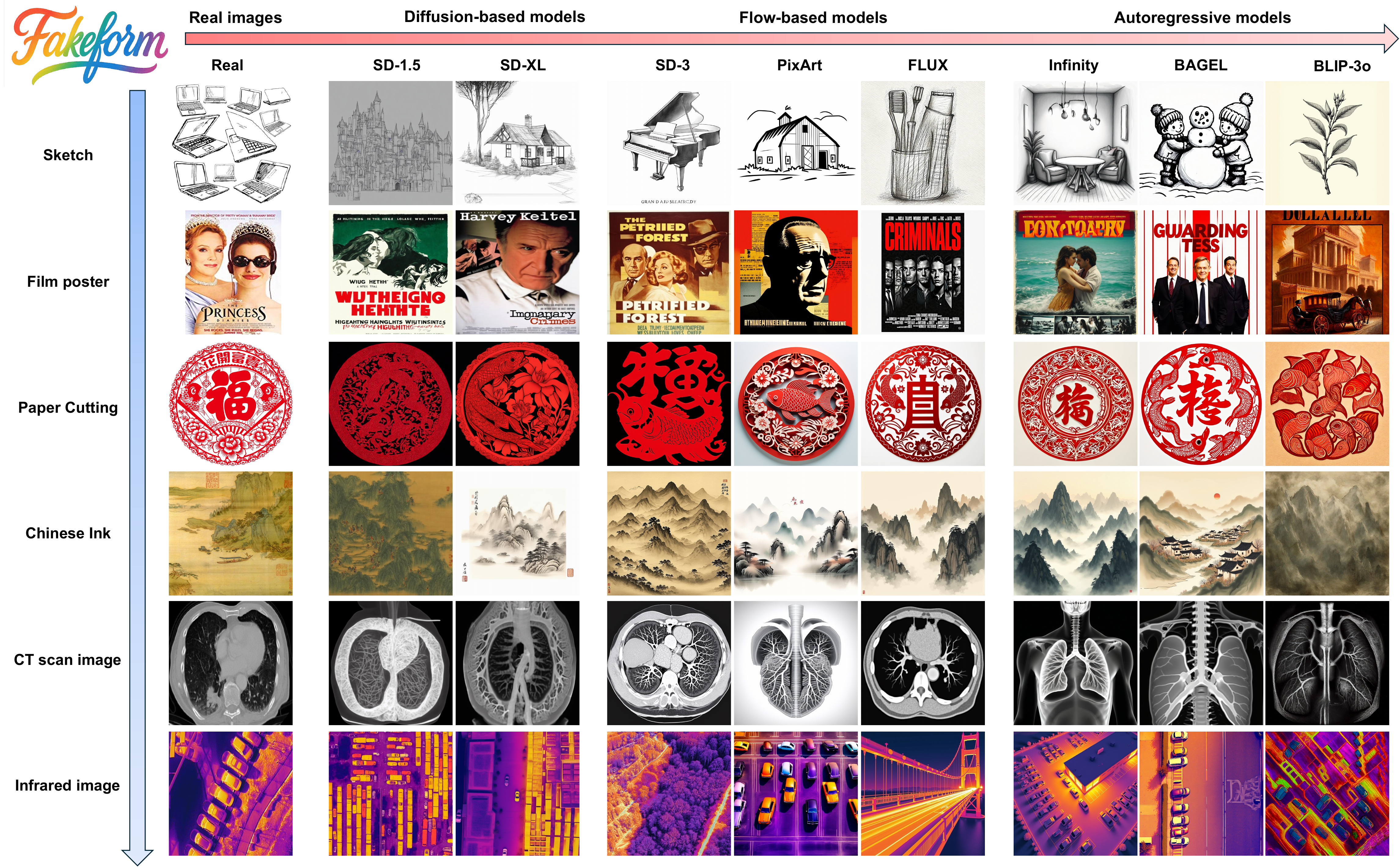}
\end{center}
\caption{Overview of the FakeForm benchmark. It features real images (left) alongside their synthetic counterparts generated by state-of-the-art models spanning diffusion, flow-matching, autoregressive, and unified multimodal paradigms. The benchmark covers domains from fine art and film to medical imaging and technical diagrams, highlighting the need for robust cross-domain generalization in AI-generated image detection.}
\label{fig:fakeform}
\end{figure*}

\begin{table*}[t]
\caption{Hierarchical structure of the 62 domains in the FakeForm benchmark.}
\label{tab:domain_framework}
\centering
\begin{minipage}{0.7\textwidth}
\renewcommand{\arraystretch}{1.0}
\setlength{\tabcolsep}{6pt}
\begin{tabular}{@{}p{0.28\linewidth} p{0.72\linewidth}@{}}
\hline
\textbf{General Photography} &
Realistic Photo, Digital Photo, ID Photo, Photojournalism \\
\hline
\textbf{Specialized Photography} &
Film Still, Astrophotography, Low-light, Fisheye, Top-down \\
\hline
\textbf{Traditional Painting} &
Paintings, Watercolor, Ink Wash, Chinese Ink Painting, Ukiyo-e \\
\hline
\textbf{Creative \& Conceptual Art} &
Concept Art, Marker Art, Poster, Murals, Graffiti, Tattoos, DeviantArt \\
\hline
\textbf{Illustration \& Comics} &
Sketch, Quickdraw, Stick Figure, Crayon, Cartoon, Comics \\
\hline
\textbf{Crafts} &
Embroidery, Origami, Paper Cutting, Reliefs \\
\hline
\textbf{Sculpture \& Objects} &
Sculptures, Wood Carving, Tile Carving, Plush, Toys, Coins \\
\hline
\textbf{Digital Graphics} &
Graphics, Game Videos, Games, UI, Emojis, Logos, Signage \\
\hline
\textbf{Scientific Imaging} &
Micrograph, CT Scan, Microorganisms, Medical Imaging \\
\hline
\textbf{Diagrams} &
Infographic, Thermodynamic Diagram, Flowcharts \\
\hline
\textbf{Data Visualization} &
Charts, Plots, Spectrograms, Traffic Maps \\
\hline
\textbf{Sensor Data} &
Depth Map, RGB--D Images, MNIST Digits \\
\hline
\textbf{Patterns} &
Patterns, Textures, Geometric Patterns \\
\hline
\textbf{Design Elements} &
Collages \\
\hline
\end{tabular}
\end{minipage}
\end{table*}

Current benchmarks for AI-generated image detection have a key limitation: they primarily evaluate generalization across generative \textit{models} within the relatively narrow domain of photorealistic images. As a result, they provide limited evidence about how detectors behave across the broader spectrum of visual content encountered in practice, such as artistic media, technical graphics, and scientific imagery.

To address this limitation, we introduce \textbf{FakeForm}, a large-scale benchmark for evaluating both cross-model and cross-domain AI-generated image detection. The FakeForm evaluation set contains approximately 370,000 images spanning 62 diverse domains, providing a substantially broader testbed than existing benchmarks. As detailed in Table~\ref{tab:domain_framework}, these domains are hierarchically organized to support both fine-grained analysis and grouped evaluation across diverse forms of visual content.

A key goal of FakeForm is to make domain generalization measurable in a controlled and practically relevant setting. The benchmark covers photographic, artistic, technical, and scientific domains, including many categories that are largely absent from previous AIGI detection datasets. This structure enables evaluation protocols in which detectors are trained on relatively standard visual content and tested on substantially different unseen domains, thereby complementing conventional cross-model benchmarks with an explicit focus on domain shift.

FakeForm also maintains broad generator coverage. It includes models from diffusion, flow-matching, autoregressive, and unified multimodal paradigms, including representative autoregressive generators such as Infinity, VAR, HART, and Hi-MAR.

The benchmark labels are defined from a provenance perspective. A sample is labeled \emph{real} when its primary visual content is authored by humans without using a generative AI model to synthesize the main image content. A sample is labeled \emph{fake} when its primary visual content is generated by a generative model. This definition applies to both photographic and non-photographic content, while hybrid or partially AI-assisted cases are outside the present scope.

A second, equally significant contribution is that, to our knowledge, FakeForm is the first benchmark in this field to be fully integrated with large-scale human perceptual data. Every test image was independently annotated by two to three experienced raters with professional backgrounds in image analysis and computer vision. Each annotation includes a 5-point perceptual realism score and a textual justification identifying the visual artifacts that influenced the assessment. To verify annotation reliability, we compute the average pairwise Spearman correlation coefficient among annotators, confirming a high degree of consistency across all domains. In total, this process yields over 760,000 human judgments, enabling a direct and quantitative comparison between machine and human detection capabilities across both model and domain axes.

Experiments on FakeForm reveal a clear practical finding: detectors that perform strongly on in-domain photorealistic data can still degrade substantially under domain shift. This highlights the importance of evaluating AIGI detection beyond conventional cross-model settings.

To support future research on robust AIGI detection, we will release FakeForm as a publicly available benchmark. The benchmark required substantial resources for data collection, generation, and annotation, and provides a rigorous testbed for studying generalization across both model and domain axes.

\section{Experiments}

\subsection{Datasets and Metrics}

\subsubsection{Prior Benchmarks}

Following established evaluation protocols, we first assess CoDA on existing public benchmarks, where detectors are typically trained on a single generator and then evaluated on diverse synthetic sources. These benchmarks provide a standard reference for cross-model generalization in photorealistic content.

\textbf{Training Dataset.}
Following prior work, we use ProGAN~\cite{karras2017progressive} as the primary training source. The training set contains 350,000 synthetic images generated by ProGAN and an equal number of real images from LSUN~\cite{yu2015lsun}, forming a balanced dataset for learning transferable forgery patterns.

\textbf{Evaluation Benchmarks.}
We evaluate on three widely used benchmarks.

\textit{ForenSynths}~\cite{wang2020cnn} focuses on generalization across traditional generators. It contains real images from six diverse sources (CelebA, FFHQ, LSUN, etc.) and synthetic images from eight generators, including StyleGAN, ProGAN, and deepfake methods. Each test subset contains 4,000 images, resulting in 48,000 total test samples.

\textit{Ojha}~\cite{ojha2023towards} evaluates performance on modern diffusion models. It contains real images from LAION and synthetic outputs from Guided Diffusion, GLIDE, LDM, and DALL-E. The benchmark further includes model variants with different inference configurations: GLIDE\_100\_10, GLIDE\_100\_27, and GLIDE\_50\_27 correspond to different denoising steps and guidance scales, while LDM\_200, LDM\_200\_cfg, and LDM\_100 test robustness to sampling settings.

\textit{GenImage}~\cite{zhu2024genimage} provides a broader evaluation across contemporary generators. It includes 1,000 ImageNet categories with real images and corresponding synthetic versions from eight generators: BigGAN, GLIDE, ADM, Midjourney, Stable Diffusion, VQDM, and Wukong. Each generator subset contains 6,000--8,000 image pairs at various resolutions, for a total of over 1.3 million test images.

\subsubsection{FakeForm Benchmark}

To evaluate broader generalization, we use FakeForm under two complementary protocols.

\textbf{Model Generalization on Photorealistic Content.}
We first evaluate cross-model generalization on the photorealistic subset of FakeForm. In this setting, all images remain within the photorealistic setting, and the goal is to measure transfer across generators rather than across domains. Following the standard setting, detectors are trained on synthetic images from Stable Diffusion v1.5 and ProGAN paired with real images, and tested on a diverse set of recent generators spanning Diffusion, Flow-Matching, Autoregressive, and Unified paradigms. The model set includes Janus-Pro, BLIP-3o, Infinity, BAGEL, VAR, HART, Hi-MAR, SD-3, PlayGround, PixArt, FLUX, SD-XL, and SD-2.0. This protocol evaluates whether a detector trained on standard photorealistic content can generalize across a broad range of contemporary image generators while remaining within the same visual domain.

\textbf{Domain Generalization Evaluation.}
Our main protocol evaluates cross-domain generalization. The training set combines 46,000 real images with 320,000 synthetic images generated by the models above, ensuring exposure to diverse generation sources during training. The cross-domain test set spans 62 diverse domains and contains approximately 370,000 images in total. Under the benchmark definition in Section~V, \emph{real} refers to human-authored provenance and \emph{fake} refers to machine-generated primary visual content. This protocol evaluates whether detectors can learn signals that transfer beyond both specific generators and familiar visual domains.

\subsubsection{Evaluation Metrics}

We report two standard metrics.

\textbf{Accuracy (Acc).}
Accuracy measures the percentage of correctly classified images using a threshold of 0.5 on the detection score.

\textbf{Average Precision (AP).}
Average Precision measures the area under the precision--recall curve across decision thresholds. Unlike Acc, which evaluates performance at a single operating point, AP provides a threshold-independent view of the precision--recall trade-off.

\begin{table*}[t]
\caption{Comparison with state-of-the-art methods on the GAN benchmark ForenSynths~\cite{wang2020cnn} (Acc/AP, \%).}
\label{tab:performance_comparison1}
\centering
\renewcommand{\arraystretch}{1.1}
\setlength{\tabcolsep}{5.5pt}
\resizebox{\textwidth}{!}{
\begin{tabular}{lccccccccc|c}
\toprule
\textbf{Method} & \textbf{Ref} & \textbf{ProGAN} & \textbf{StyleGAN} & \textbf{StyleGAN2} & \textbf{BigGAN} & \textbf{CycleGAN} & \textbf{StarGAN} & \textbf{GauGAN} & \textbf{Deepfake} & \textbf{Mean} \\
\midrule
CNNDet & CVPR 2020 \cite{wang2020cnn} & 91.4 / 99.2 & 63.8 / 91.4 & 76.4 / 96.7 & 52.9 / 73.3 & 72.6 / 88.2 & 63.8 / 90.7 & 63.8 / 92.2 & 51.7 / 62.3 & 76.0 / 86.7 \\
FreDect & ICML 2020 \cite{frank2020leveraging} & 90.3 / 85.2 & 74.4 / 72.2 & 73.1 / 71.4 & 88.7 / 86.5 & 75.4 / 71.7 & 99.5 / 99.5 & 69.0 / 77.4 & 60.7 / 49.2 & 78.9 / 76.6 \\
LGrad & CVPR 2023 \cite{tan2023learning} & 99.8 / 100.0 & 94.8 / 99.9 & 96.0 / 99.9 & 82.9 / 90.5 & 85.3 / 94.7 & 99.6 / 100.0 & 72.4 / 79.2 & 58.1 / 67.8 & 86.1 / 91.6 \\
UFD & CVPR 2023 \cite{ojha2023towards} & 99.8 / 100.0 & 89.0 / 98.8 & 83.8 / 98.6 & 90.5 / 99.1 & 87.8 / 99.6 & 91.2 / 100.0 & 89.9 / 99.2 & 80.1 / 90.2 & 89.1 / 98.2 \\
PatchCraft & arXiv 2023 \cite{zhong2023rich} & \best{100.0 / 100.0} & 93.0 / 98.7 & 89.7 / 97.7 & \secondbest{95.7 / 99.3} & 70.0 / 85.1 & \best{100.0 / 100.0} & 71.9 / 81.8 & 58.6 / 79.6 & 84.8 / 92.7 \\
FreqNet & AAAI 2024 \cite{tan2024frequency} & 99.6 / 100.0 & 90.2 / 99.6 & 88.0 / 95.5 & 90.5 / 95.5 & 95.5 / 99.7 & 85.8 / 100.0 & 93.4 / 98.6 & 88.9 / 94.5 & 91.5 / 97.9 \\
NPR & CVPR 2024 \cite{tan2024rethinking} & 99.8 / 100.0 & 96.3 / 100.0 & 97.2 / 100.0 & 87.5 / 94.5 & 95.0 / 95.7 & \secondbest{99.7 / 100.0} & 86.6 / 88.2 & 77.4 / 86.1 & 92.4 / 95.6 \\
FatFormer & CVPR 2024 \cite{liu2024forgery} & 99.7 / 100.0 & 97.2 / 99.6 & \best{98.7 / 99.8} & \best{99.5 / 100.0} & \secondbest{99.2 / 99.8} & \secondbest{99.7 / 100.0} & \secondbest{99.3 / 100.0} & \secondbest{93.2 / 97.6} & 98.2 / 99.5 \\
SAFE & KDD 2025 \cite{li2024improving} & \secondbest{99.9 / 100.0} & \secondbest{97.8 / 99.8} & \secondbest{98.6 / 100.0} & 89.5 / 95.4 & \best{99.6 / 99.8} & \secondbest{99.7 / 100.0} & 91.4 / 97.0 & 91.4 / 97.5 & \secondbest{96.0 / 98.7} \\
\midrule
\textbf{CoDA} & Ours & \secondbest{99.9 / 100.0} & \best{98.2 / 99.9} & \secondbest{98.6 / 99.9} & 93.6 / 98.0 & 99.1 / 99.9 & \best{100.0 / 100.0} & \best{99.9 / 99.9} & \best{93.4 / 98.5} & \best{98.2 / 99.6} \\
\bottomrule
\end{tabular}}
\end{table*}

\begin{table*}[t]
\caption{Comparison with state-of-the-art methods on the diffusion benchmark of Ojha et al.~\cite{ojha2023towards} (Acc/AP, \%).}
\label{tab:performance_comparison2}
\centering
\renewcommand{\arraystretch}{1.1}
\setlength{\tabcolsep}{5.5pt}
\resizebox{\textwidth}{!}{
\begin{tabular}{lccccccccc|c}
\toprule
\textbf{Method} & \textbf{Ref} & \textbf{DALL-E} & \textbf{GLIDE\_100\_10} & \textbf{GLIDE\_100\_27} & \textbf{GLIDE\_50\_27} & \textbf{ADM} & \textbf{LDM\_100} & \textbf{LDM\_200} & \textbf{LDM\_200\_cfg} & \textbf{Mean} \\
\midrule
CNNDet & CVPR 2020 \cite{wang2020cnn}  & 51.8 / 61.2 & 53.3 / 72.9 & 53.1 / 71.3 & 54.2 / 76.1 & 54.9 / 66.6 & 51.9 / 63.7 & 52.0 / 64.5 & 51.6 / 63.1 & 52.8 / 67.5 \\
FreDect & ICML 2020 \cite{frank2020leveraging} & 57.0 / 62.5 & 53.6 / 44.3 & 50.4 / 40.8 & 52.0 / 42.3 & 53.4 / 52.5 & 56.6 / 51.3 & 56.4 / 50.9 & 56.4 / 52.4 & 54.5 / 49.6 \\
LGrad & CVPR 2023 \cite{tan2023learning} & 88.5 / 97.3 & 89.5 / 94.9  & 87.4 / 93.2 & 90.6 / 95.0 & \secondbest{86.7 / 99.8} & 94.8 / 99.2 & 94.2 / 99.1 & 95.8 / 99.2 &  90.9 / 97.3 \\
UFD & CVPR 2023 \cite{ojha2023towards} & 89.3 / 96.5 & 90.0 / 96.5 & 90.7 / 97.0 & 90.8 / 97.2 & 75.1 / 84.5 & 90.1 / 97.0 & 90.2 / 97.0 & 77.3 / 88.6 & 86.7 / 94.3 \\
PatchCraft & arXiv 2023 \cite{zhong2023rich} & 83.3 / 93.0 & 80.1 / 92.0 & 83.4 / 93.9 & 77.6 / 88.7 & 80.9 / 90.5 & 88.9 / 97.7 & 89.3 / 97.9 & 88.1 / 96.9 & 84.0 / 93.8 \\
FreqNet & AAAI 2024 \cite{tan2024frequency} & 97.2 / 99.5 & 87.8 / 96.1 & 84.4 / 96.6 & 86.5 / 95.0 & 67.2 / 74.5 & 97.8 / 99.6 & 97.4 / 99.0 & 98.2 / 99.0 & 89.6 / 94.9 \\
NPR & CVPR 2024 \cite{tan2024rethinking} & 94.5 / 99.5 & \best{98.2 / 99.8} & \best{97.8 / 99.7} & \best{98.2 / 99.8} & 75.8 / 81.0 & \best{99.3 / 99.0} & \best{99.1 / 99.9} & \best{99.0 / 99.9} & 95.1 / 97.4 \\
FatFormer & CVPR 2024 \cite{liu2024forgery} & \best{98.7 / 99.8} & 94.5 / 99.5 & 94.1 / 99.3 & 94.2 / 99.1 & 75.9 / 91.8 & 98.5 / 99.8 & 98.5 / 99.8 & 94.8 / 99.0 & 93.6 / 98.4 \\
SAFE & KDD 2025 \cite{li2024improving} & 97.5 / 99.7 & 97.2 / 99.4 & 95.8 / 98.9 & 96.6 / 99.2 & 82.4 / 95.7 & \secondbest{98.8 / 100.0} &  \secondbest{98.8 / 100.0} & \secondbest{98.7 / 99.8} & \secondbest{95.7 / 99.0} \\
\midrule
\textbf{CoDA} & Ours & \secondbest{98.1 / 99.6} & \secondbest{97.6 / 99.6} & \secondbest{97.4 / 99.5} & \secondbest{97.7 / 99.5} & \best{92.8 / 97.4} & \secondbest{98.8 / 99.8} & \secondbest{98.9 / 100.0} & \secondbest{98.7 / 99.8} & \best{97.5 / 99.4} \\
\bottomrule
\end{tabular}}
\end{table*}

\begin{table*}[t]
\caption{Comparison with state-of-the-art methods on GenImage~\cite{zhu2024genimage} (Acc/AP, \%).}
\label{tab:performance_comparison3}
\centering
\renewcommand{\arraystretch}{1.1}
\setlength{\tabcolsep}{5.5pt}
\resizebox{\textwidth}{!}{
\begin{tabular}{lccccccccc|c}
\toprule
\textbf{Method} & \textbf{Ref} & \textbf{Midjourney} & \textbf{SD v1.4} & \textbf{SD v1.5} & \textbf{ADM} & \textbf{GLIDE} & \textbf{Wukong} & \textbf{VQDM} & \textbf{BigGAN} & \textbf{Mean} \\
\midrule
CNNDet & CVPR 2020 \cite{wang2020cnn}  & 50.1 / 53.4 & 50.3 / 55.9 & 50.3 / 56.1 & 53.0 / 69.2 & 51.7 / 66.9 & 51.4 / 62.4 & 50.0 / 53.5 & 69.8 / 91.5 & 53.3 / 63.6 \\
FreDect & ICML 2020 \cite{frank2020leveraging} & 32.1 / 35.7 & 28.8 / 34.9 & 28.9 / 34.6 & 62.9 / 70.1 & 42.8 / 42.2 & 35.9 / 38.0 & 72.1 / 84.2 & 26.1 / 34.7 & 41.2 / 46.7 \\
LGrad & CVPR 2023 \cite{tan2023learning} & 73.7 / 77.5 & 76.3 / 80.1 & 77.4 / 80.1 & 51.8 / 51.0 & 49.8 / 50.5 & 73.1 / 75.4 & 52.1 / 51.5 & 40.5 / 30.2 & 61.8 / 62.0 \\
UFD & CVPR 2023 \cite{ojha2023towards} & 56.9 / 68.5 & 65.1 / 81.5 & 64.7 / 81.0 & 69.2 / 84.2 & 60.1 / 73.5 & 73.5 / 89.0 & 86.0 / 95.0 & 89.3 / 97.0 & 70.6 / 83.7 \\
PatchCraft & arXiv 2023 \cite{zhong2023rich} & 89.7 / 96.2 & 95.0 / 98.9 & 94.6 / 98.8 & 81.6 / 93.3 & 83.5 / 93.8 & 90.9 / 97.4 & 88.2 / 95.9 & 91.5 / 97.8 & 86.3 / 96.4 \\
FreqNet & AAAI 2024 \cite{tan2024frequency} & 69.7 / 78.5 & 64.2 / 74.5 & 64.9 / 75.6 & 83.5 / 92.0 & 81.2 / 88.5 & 57.8 / 67.0 & 81.4 / 90.0 & 90.5 / 95.0 & 74.1 / 82.6 \\
NPR & CVPR 2024 \cite{tan2024rethinking} & 77.8 / 85.4 & 78.6 / 84.0 & 78.9 / 84.6 & 69.7 / 74.6 & 78.4 / 85.7 & 76.1 / 80.5 & 78.1 / 81.0 & 80.1 / 88.2 & 77.2 / 83.0 \\
FatFormer & CVPR 2024 \cite{liu2024forgery} & 56.0 / 62.7 & 67.8 / 81.1 & 67.3 / 81.4 & 78.2 / 91.5 & 87.9 / 95.5 & 73.0 / 85.7 & 86.8 / 96.9 & 96.7 / 99.0 & 76.7 / 86.7 \\
SAFE & KDD 2025 \cite{li2024improving} & \secondbest{95.2 / 99.0} & \best{99.4 / 99.1} & \secondbest{99.3 / 99.7} & 82.2 / 96.7 & \best{96.2 / 99.3} & \secondbest{98.1 / 99.8} & 96.2 / 99.4 & \best{97.7 / 99.8} & \secondbest{95.5 / 99.1}\\
AIDE $\dagger$ & ICLR 2025 \cite{yan2024sanity} & 77.2 / 83.0 & 93.0 / 97.0 & 92.8 / 97.0 & \best{93.4 / 97.0} & 95.1 / 98.0 & 93.6 / 98.0 & \secondbest{96.6 / 99.0} & 84.0 / 94.0 & 92.8 / 95.4 \\
AIGI-Holmes $\dagger$ & ICCV 2025 \cite{zhou2025aigi} & 81.6 / 87.0 & 91.3 / 96.0 & 91.4 / 96.0 & 91.5 / 96.0 & 88.4 / 93.0 & 89.5 / 95.0 & 90.9 / 96.0 & 94.5 / 98.0 & 93.2 / 94.6 \\
\midrule
\textbf{CoDA} & Ours & \best{96.0 / 99.0} & \secondbest{99.2 / 99.0} & \best{99.8 / 100.0} & \secondbest{85.2 / 96.8} & \secondbest{95.9 / 99.2} & \best{97.8 / 99.6} & \best{96.8 / 99.9} & \secondbest{96.3 / 98.2} & \best{95.9 / 99.1} \\
\bottomrule
\end{tabular}}
\begin{minipage}{0.98\textwidth}
\footnotesize
\vspace{2pt}
\emph{Note:} Results in this table are averaged over three independent training and evaluation runs. The corresponding result reported in the CVPR version~\cite{jia2025secret} was based on a single run, which accounts for the minor numerical difference.
\end{minipage}
\end{table*}

\subsection{Implementation Details}

CoDA uses a lightweight ResNet backbone with only 1.48M parameters. The architecture processes both the original image and the difference map produced by the Noise-Quantization Probe. We train the model using AdamW with an initial learning rate of $10^{-4}$, batch size 512, and weight decay 0.01 for 200 epochs. The learning rate follows a cosine annealing schedule. For the probe, we set the noise standard deviation to $\sigma=0.10$ and use $R=50$ replicas according to validation performance. Images are randomly cropped to $256\times256$ during training and center-cropped during testing. We apply ColorJitter (brightness, contrast, saturation factor 0.2), random rotation ($\pm90^\circ$), and random occlusion (50\% probability, 25--50\% mask ratio). For difference maps, we apply geometric augmentations such as crop and flip to preserve the probe signal. All experiments are implemented in PyTorch and conducted on a single NVIDIA A100 GPU.

For fair comparison, we divide baseline methods into two groups. Models marked with $\dagger$ only provide inference APIs or are part of large closed-source multimodal systems; these models are evaluated directly through their released services without retraining. All other baselines are reproduced using their official implementations and recommended settings. Unless otherwise specified, all reported quantitative results are averaged over three independent runs.

\subsection{Performance on Existing Benchmarks}

We begin with established public benchmarks to verify that CoDA remains competitive under standard cross-model evaluation. These experiments provide a conventional reference point, while the broader cross-domain analysis is deferred to FakeForm.

\textbf{Robustness to Diverse GANs.}
On the GAN-centric ForenSynths benchmark~\cite{wang2020cnn}, CoDA achieves state-of-the-art performance with 98.2/99.6 Acc/AP (Table~\ref{tab:performance_comparison1}), outperforming or matching prior leading methods such as FatFormer. CoDA also remains stable across varied GAN architectures, from earlier models such as CycleGAN to stronger generators such as StyleGAN2, suggesting that the learned cue is not tied to a specific adversarial architecture.

\textbf{Generalization from GANs to Diffusion Models.}
A key test for modern detectors is generalizing from GAN-based training to the distinct artifacts of diffusion models. On the diffusion-only dataset from Ojha et al.~\cite{ojha2023towards}, CoDA achieves the best mean result with 97.5\% accuracy and 99.4\% AP (Table~\ref{tab:performance_comparison2}). 

This advantage carries over to the highly diverse GenImage benchmark~\cite{zhu2024genimage}, where CoDA again achieves the best overall performance (95.9\% accuracy, 99.1\% AP; Table~\ref{tab:performance_comparison3}), outperforming competitors on most recent models, including Midjourney and Stable Diffusion variants. In contrast, methods relying on pre-trained semantic features (e.g., UFD, FatFormer) or frequency analysis (FreqNet) exhibit marked degradation when moving from GANs to diffusion models, whereas CoDA remains consistently strong. These results suggest that the probe-induced color-distribution cue captured by CoDA transfers more consistently across generator families than cues tied primarily to high-level semantics or specific frequency patterns.

\subsection{Performance on FakeForm}

Traditional benchmarks are essential for assessing cross-model generalization, but they are largely confined to photorealistic imagery. We next report results on FakeForm, which enables evaluation of both cross-model and cross-domain generalization in a broader setting.

\textbf{Cross-Model Generalization.}
We first consider cross-model generalization on the photorealistic subset of FakeForm. In this setting, all test images belong to the photorealistic setting, and the goal is to measure transfer across generators rather than across domains. As reported in Table~\ref{tab:fakeform_model_split_new}, the evaluated generators span Diffusion, Flow-Matching, Unified, and Autoregressive paradigms, including Janus-Pro, BLIP-3o, Infinity, BAGEL, VAR, HART, Hi-MAR, SD-3, PlayGround, PixArt, FLUX, SD-XL, and SD-2.0. CoDA achieves strong overall performance, surpassing most specialized detectors and remaining highly competitive with large VLM-based systems. It also attains top-tier results on several recent generators, including Infinity, PlayGround, SD-XL, and SD-2.0. These results indicate that the proposed cue transfers well across diverse generation paradigms within the photorealistic setting.

\textbf{Cross-Domain Generalization.}
We further evaluate domain generalization by training on photorealistic images and testing across 62 diverse visual domains. As shown in Table~\ref{tab:fakeform_domain_split_revised_v3}, CoDA achieves the highest mean performance of 77.7/88.1 Acc/AP, outperforming all compared methods, including large VLMs. Methods that perform strongly on photorealistic content, such as FatFormer (76.5/87.2) and AIGI-Holmes (73.6/86.4), degrade more substantially under domain shift, whereas CoDA remains comparatively stable. Its advantage is particularly clear in domains far from natural photography, such as \emph{Paper Cutting}, \emph{Depth Map}, and \emph{Film Still}. At the same time, performance remains domain-dependent, with lower results on domains such as \emph{Sketch} and \emph{Thermodynamic Diagram}. This observation motivates the more detailed domain-wise analyses and failure-case discussion that follow.

\begin{table*}[!t]
\caption{Comparison on FakeForm with respect to different \textbf{generative models} (Acc/AP, \%). The model set spans diffusion, flow-matching, unified, and autoregressive paradigms.}
\label{tab:fakeform_model_split_new}
\centering
\renewcommand{\arraystretch}{1.2}
\setlength{\tabcolsep}{4pt}
\resizebox{\textwidth}{!}{
\begin{tabular}{l|cccccccccccccc|c}
\toprule
\textbf{Method} &
\textbf{Janus-Pro} & \textbf{BLIP-3o} & \textbf{Infinity} & \textbf{BAGEL} &
\textbf{VAR} & \textbf{HART} & \textbf{Hi-MAR} & \textbf{SD-3} &
\textbf{PlayGround} & \textbf{PixArt} & \textbf{FLUX} & \textbf{SD-XL} &
\textbf{SD-2.0} & \textbf{Mean} \\
\midrule
\textbf{Humans} & 72.5 / 82.1 & 74.2 / 82.4 & 79.6 / 87.0 & 78.2 / 86.1 & 73.5 / 84.7 & 72.5 / 82.8 & 74.0 / 84.0 & 69.4 / 78.6 & 75.4 / 81.9 & 69.3 / 74.5 & 79.2 / 89.7 & 75.3 / 83.9 & 86.8 / 91.7 & 75.4 / 83.8 \\
CNNDet & 63.8 / 79.0 & 63.5 / 78.4 & 77.7 / 86.0 & 66.1 / 79.3 & 69.5 / 81.5 & 68.3 / 79.5 & 70.8 / 80.2 & 55.5 / 65.0 & 79.5 / 86.9 & 52.8 / 71.5 & 70.6 / 83.7 & 56.9 / 74.3 & 72.8 / 85.2 & 66.8 / 79.3 \\
UFD & 65.9 / 80.1 & 76.6 / 86.0 & 77.2 / 85.5 & 79.0 / 88.0 & 82.5 / 89.1 & 76.0 / 84.3 & 78.2 / 85.8 & 88.9 / 91.9 & 76.5 / 87.0 & 83.2 / 88.1 & 74.1 / 85.3 & 85.4 / 90.6 & 59.7 / 74.0 & 77.2 / 85.8 \\
FreqNet & 80.7 / 90.5 & 89.5 / 91.7 & 89.0 / 91.6 & 79.0 / 89.8 & 83.8 / 90.9 & 81.5 / 87.2 & 83.5 / 88.6 & 87.2 / 91.7 & 72.2 / 88.8 & 57.7 / 77.4 & 68.8 / 86.2 & 80.1 / 90.6 & 63.3 / 84.3 & 78.2 / 88.4 \\
NPR & 45.1 / 49.4 & 62.1 / 67.7 & 66.2 / 70.5 & 86.4 / 92.9 & 69.4 / 83.2 & 64.0 / 71.8 & 65.8 / 73.5 & 86.6 / 92.9 & 87.2 / 93.1 & 78.3 / 83.9 & 85.9 / 92.4 & 89.4 / 90.7 & 86.1 / 92.5 & 74.8 / 81.1 \\
FatFormer & 64.1 / 92.0 & 67.1 / 92.6 & 88.2 / 91.6 & 92.9 / 93.1 & 81.5 / 92.0 & 80.8 / 87.5 & 82.5 / 88.8 & \secondbest{93.0 / 94.5} & 86.8 / 92.5 & 74.3 / 90.6 & 91.6 / 93.0 & 89.1 / 91.6 & 88.0 / 92.5 & 83.1 / 91.7 \\
SAFE & 82.6 / 91.9 & 84.4 / 92.2 & 92.1 / 93.3 & 92.0 / 93.9 & 87.5 / 93.0 & 85.2 / 89.0 & 87.0 / 90.5 & \best{93.2 / 93.9} & 90.9 / 92.2 & 77.4 / 90.1 & 91.7 / 92.4 & \best{92.6 / 93.8} & \secondbest{93.1 / 93.8} & 88.4 / 92.3 \\
AIDE $\dagger$ & 84.0 / 92.0 & \best{91.2 / 93.3} & \secondbest{92.5 / 93.2} & \best{94.3 / 93.5} & \secondbest{90.8 / 93.6} & \best{89.5 / 91.8} & 90.2 / 91.5 & 92.6 / 93.7 & \secondbest{92.4 / 93.5} & \best{91.8 / 93.7} & \best{92.7 / 93.8} & 86.5 / 88.0 & 92.0 / 92.8 & 90.8 / 92.6 \\
AIGI-Holmes $\dagger$ & \best{90.4 / 93.0} & \secondbest{91.5 / 93.0} & 91.1 / 92.9 & 93.0 / 93.1 & \best{92.2 / 93.0} & 89.2 / 91.2 & \best{91.0 / 92.0} & 93.2 / 93.1 & 93.0 / 93.1 & 91.3 / 92.9 & \secondbest{92.6 / 93.0} & 93.1 / 92.9 & 91.7 / 93.0 & \best{91.8 / 92.8} \\
\midrule
\textbf{CoDA (Ours)} & \secondbest{86.2 / 92.4} & 88.3 / 91.8 & \best{93.1 / 94.9} & \secondbest{92.6 / 93.8} & 91.2 / 92.3 & 87.8 / 90.5 & 89.5 / 91.2 & 91.2 / 93.6 & \best{93.8 / 94.6} & \secondbest{91.7 / 92.9} & 90.4 / 93.2 & \secondbest{93.2 / 93.4} & \best{93.7 / 94.4} & \secondbest{91.0 / 93.0} \\
\bottomrule
\end{tabular}}
\end{table*}

\begin{table*}[!t]
\caption{Comparison on representative FakeForm \textbf{visual domains} and the mean over all 62 domains (Acc/AP, \%). Humans are reported as a reference and are not considered when marking the best and second-best detector results.}
\label{tab:fakeform_domain_split_revised_v3}
\centering
\renewcommand{\arraystretch}{1.2}
\setlength{\tabcolsep}{4pt}
\resizebox{\textwidth}{!}{
\begin{tabular}{l| ccccccccccc c}
\toprule
\textbf{Method} &
\textbf{Photojournalism} & \textbf{Sketch} & \textbf{Quickdraw} & \textbf{Poster} &
\textbf{Watercolor} & \textbf{Concept Art} & \textbf{Stick Figure} &
\textbf{Ukiyo-e} & \textbf{Paper Cutting} & \textbf{Crayon} &
\textbf{Cartoon} & \textbf{Chinese Ink Painting} \\
\midrule
\textbf{Humans}     & 79.3/87.9 & 65.5/74.1 & 55.4/64.2 & 84.8/92.3 & 60.7/71.5 & 74.9/81.6 & 61.5/71.0 & 70.5/80.3 & 68.3/75.3 & 65.3/73.5 & 70.5/79.3 & 71.5/82.3 \\
CNNDet   & 55.4/60.1 & 50.3/53.2 & 54.2/67.8 & 53.5/45.0 & 52.5/55.3 & 43.3/46.5 & 53.1/57.5 & 54.2/53.5 & 50.3/65.5 & 56.2/59.8 & 49.3/60.5 & 49.4/51.2 \\
UFD        & 73.2/77.1 & 67.8/75.3 & 57.3/70.5 & 78.3/86.1 & \secondbest{71.3/78.5} & \best{75.3/82.1} & \best{69.0/81.5} & 73.2/87.2 & 51.5/67.7 & 73.8/81.7 & 71.3/81.3 & 75.3/84.5 \\
FreqNet    & 58.1/62.3 & 58.1/65.2 & 82.3/97.7 & 55.4/47.2 & 55.1/57.8 & 47.0/47.3 & 53.9/58.8 & 60.5/55.7 & 62.0/70.8 & 58.3/61.6 & 62.4/83.5 & 50.5/52.0 \\
NPR        & 70.5/83.4 & 58.5/66.7 & \secondbest{91.3/96.6} & 63.1/79.5 & 65.5/78.4 & 44.5/48.7 & 68.1/77.5 & 81.5/80.7 & 82.5/88.8 & 71.5/80.2 & 72.0/82.2 & 71.4/86.5 \\
FatFormer  & \secondbest{83.4/91.5} & \best{75.8/85.3} & 55.5/95.3 & \secondbest{93.1/97.8} & 67.3/82.0 & 60.2/81.1 & 68.5/77.5 & \secondbest{82.3/93.4} & 87.8/95.9 & \secondbest{78.5/90.7} & \secondbest{78.0/88.9} & \best{77.3/90.4} \\
SAFE       & 63.5/65.8 & 52.0/54.5 & 89.1/96.5 & 62.3/70.0 & 64.8/79.5 & 57.1/57.5 & 58.2/58.8 & 69.4/73.5 & 85.5/92.3 & 59.3/63.4 & 50.5/61.5 & 64.0/61.3 \\
AIDE $\dagger$ & 62.0/79.6 & \secondbest{68.0/78.2} & \best{97.3/99.8} & 63.1/68.7 & 65.2/82.2 & \secondbest{67.5/81.5} & 66.4/84.2 & 55.6/88.4 & \secondbest{89.8/96.3} & 61.4/81.8 & 62.0/80.5 & 59.4/90.8 \\
AIGI-Holmes $\dagger$ & 80.3/86.7 & 72.5/80.2 & 80.3/85.5 & 89.3/89.2 & 64.8/75.3 & 58.5/70.2 & 65.0/78.8 & 78.5/85.3 & 82.3/88.7 & 75.2/82.5 & \best{79.2/91.2} & \secondbest{72.8/84.0} \\
\textbf{CoDA (Ours)} & \best{87.3/94.0} & 65.1/75.9 & 58.4/76.8 & \best{94.5/98.6} & \best{75.5/88.3} & 62.3/81.8 & \secondbest{68.3/79.8} & \best{87.5/95.7} & \best{92.5/97.5} & \best{80.3/92.0} & 75.3/87.7 & 70.4/82.0 \\
\midrule
& \textbf{Infographic} & \textbf{Mural} & \textbf{Micrograph} & \textbf{CT Scan} &
\textbf{Depth Map} & \textbf{Film Still} & \textbf{Wood Carving} &
\textbf{Tile Carving} & \textbf{Thermodyn. Diagram} & \textbf{Game Videos} & \textbf{Coin} &
\textbf{Mean (62 Domains)} \\
\midrule
\textbf{Humans}     & 88.3/93.9 & 78.0/85.1 & 60.3/72.6 & 75.4/88.3 & 65.3/75.3 & 78.7/86.5 & 85.2/92.6 & 82.4/90.3 & 71.3/82.9 & 70.3/78.5 & 80.5/88.3 & 72.1/81.6 \\
CNNDet     & 55.8/71.3 & 40.3/38.7 & 38.3/37.4 & 75.4/88.3 & 20.3/45.3 & 58.7/60.5 & 61.5/71.0 & 53.2/62.1 & 31.3/35.3 & 42.3/48.5 & 50.5/50.3 & 50.4/52.3 \\
UFD        & 85.5/95.0 & \best{89.8/97.3} & \best{87.8/97.9} & 76.3/89.5 & 43.3/47.5 & 80.8/90.5 & 74.2/87.0 & 74.3/93.1 & 69.2/86.4 & 64.2/76.3 & 67.2/81.6 & 71.2/83.8 \\
FreqNet    & 61.8/74.7 & 42.2/40.2 & 57.8/73.5 & 94.8/99.2 & 76.3/93.2 & 63.4/64.7 & 71.3/81.3 & 74.2/86.3 & 33.4/37.1 & 43.4/49.5 & 66.0/74.0 & 61.3/66.3 \\
NPR        & 65.4/71.0 & 78.3/88.7 & 78.7/89.1 & 95.5/98.5 & 80.1/94.1 & 63.4/78.0 & 67.3/84.1 & 60.4/85.2 & 43.2/61.4 & 64.5/84.1 & 63.1/78.1 & 68.4/82.1 \\
FatFormer  & \secondbest{81.9/94.3} & 82.5/93.3 & 69.4/83.4 & \best{98.6/99.8} & \secondbest{91.2/96.8} & \secondbest{90.8/97.3} & 84.8/92.9 & \secondbest{84.3/94.0} & 56.1/80.9 & 74.4/90.2 & \secondbest{74.1/90.8} & \secondbest{76.5/87.2} \\
SAFE       & 61.4/72.8 & 70.0/88.1 & 39.5/38.1 & 89.2/96.5 & 21.5/59.7 & 60.5/62.3 & 63.8/73.2 & 54.4/63.7 & 50.2/83.5 & 53.4/72.7 & 52.1/52.2 & 58.6/67.1 \\
AIDE $\dagger$ & 67.3/78.9 & 53.4/76.2 & 55.2/74.7 & 83.4/93.6 & 68.5/92.4 & 65.4/85.7 & 61.1/90.0 & 58.3/85.5 & \secondbest{58.5/82.8} & 64.5/79.0 & 63.4/84.7 & 67.0/83.0 \\
AIGI-Holmes $\dagger$ & 78.3/84.9 & 79.3/86.7 & 65.5/80.4 & \secondbest{94.8/96.3} & 85.4/91.8 & 87.9/93.3 & \secondbest{84.8/87.9} & 82.4/90.3 & 51.5/75.7 & \best{78.4/86.2} & 70.2/88.5 & 73.6/86.4 \\
\textbf{CoDA (Ours)} & 65.4/74.5 & \secondbest{87.9/95.0} & \secondbest{79.5/90.3} & 92.3/94.8 & \best{92.8/97.7} & \best{94.5/98.6} & 70.3/88.0 & 71.3/88.3 & 55.4/72.6 & \secondbest{77.1/86.3} & 75.5/88.3 & \best{77.7/88.1} \\
\bottomrule
\end{tabular}}
\end{table*}

\subsection{Efficiency}

CoDA also provides a favorable balance between robustness and efficiency. Many recent detectors rely on large VLMs to achieve strong generalization, but their computational cost limits practical deployment. In contrast, CoDA achieves high accuracy with real-time throughput using a compact architecture that explicitly exploits probe-induced color cues. As summarized in Table~\ref{tab:efficiency_comparison}, CoDA processes 125.2 frames per second, whereas VLM-based systems such as AIGI-Holmes reach only 2.2 FPS. CoDA is also extremely compact, with only 1.48M parameters, compared with the 7B+ scale of AIGI-Holmes.

Crucially, this efficiency does not come at the expense of robustness. Compared with other lightweight detectors such as FreqNet and SAFE, CoDA maintains stronger performance under both generator and domain shifts. This makes it a practical solution for settings that require both scalability and broad generalization.

\begin{table}[!t]
\caption{Efficiency comparison of detectors. All results are measured on a single NVIDIA A100 40G GPU. FLOPs are unavailable for closed-source/API systems.}
\label{tab:efficiency_comparison}
\centering
\renewcommand{\arraystretch}{1.2}
\setlength{\tabcolsep}{5pt}
\begin{tabular}{lcccc}
\toprule
\textbf{Method} & \textbf{Backbone} & \textbf{FPS}~$\uparrow$ & \textbf{FLOPs (G)}~$\downarrow$ & \textbf{Params (M)}~$\downarrow$ \\
\midrule
\multicolumn{5}{c}{\textit{Non-real-time detection ($\leq$30 FPS)}} \\
\midrule
CNNDet                   & Xception & 26.2  & 16.57  & 23.51 \\
FatFormer                & ViT      & 13.3  & 216.83 & 492.59 \\
AIDE                     & ResNeXt  & 18.9  & 225.69 & 893.54 \\
AIGI-Holmes              & ViT      & 2.2   & --     & $>7000$ \\
\midrule
\multicolumn{5}{c}{\textit{Real-time detection ($>$30 FPS)}} \\
\midrule
PatchCraft               & ResNet   & 31.9  & 4.13   & 0.19 \\
FreqNet                  & ResNet   & 157.2 & 1.97   & 1.85 \\
NPR                      & ResNet   & 200.6 & 1.75   & 1.44 \\
SAFE                     & ResNet   & 180.4 & 2.29   & 1.44 \\
\textbf{CoDA (Ours)}      & ResNet   & 125.2 & 2.37   & 1.48 \\
\bottomrule
\end{tabular}
\end{table}

\subsection{Robustness to Image Perturbations}

To assess practical robustness, we evaluate CoDA under common image perturbations that simulate real-world degradation during storage and transmission. Following the protocol of FatFormer~\cite{liu2024forgery}, we apply four transformations with 50\% probability: Gaussian blurring, random cropping, JPEG compression (QF=75), and Gaussian noise. The evaluation is conducted on the GAN-based ForenSynths and the diffusion-based GenImage test sets.

The results in Table~\ref{tab:robustness_wide} show that CoDA remains highly resilient under these perturbations. The advantage is especially clear on diffusion-generated images, where competing methods suffer larger performance drops. For example, under JPEG compression, CoDA decreases only from 95.3\% to 93.7\% in Acc, whereas other methods degrade more substantially. This behavior is consistent with the proposed probe mechanism: perturbations such as blurring and compression may weaken high-frequency or semantic cues, but the broader statistical regularities captured by the Noise-Quantization Probe remain relatively stable.

\begin{table*}[!t]
\centering
\caption{Robustness evaluation against common image perturbations (Acc/AP, \%).}
\label{tab:robustness_wide}
\renewcommand{\arraystretch}{1.1}
\setlength{\tabcolsep}{8pt}
\begin{tabular}{@{}ll ccccc@{}}
\toprule
\textbf{Method} & \textbf{Dataset} & \textbf{Original} & \textbf{Gaussian Blurring} & \textbf{Random Cropping} & \textbf{JPEG (QF=75)} & \textbf{Gaussian Noise} \\
\midrule
\multirow{2}{*}{LGrad} & GANs & 86.1 / 91.6 & 78.5 / 83.2 & 83.5 / 90.8 & 77.4 / 85.2 & 76.1 / 84.4 \\
& Diffusion & 61.8 / 62.0 & 57.3 / 58.4 & 53.1 / 52.4 & 58.8 / 59.9 & 56.7 / 57.8 \\
\midrule
\multirow{2}{*}{UFD} & GANs & 89.1 / 98.2 & 83.4 / 92.0 & 84.8 / 94.1 & 85.5 / 93.7 & 81.3 / 92.6 \\
& Diffusion & 68.3 / 78.4 & 68.3 / 78.4 & 62.9 / 72.0 & 64.8 / 76.7 & 69.2 / 80.1 \\
\midrule
\multirow{2}{*}{FatFormer} & GANs & \textbf{98.2} / 99.5 & 97.7 / 98.4 & 97.3 / \textbf{98.8} & 96.7 / 99.1 & 94.4 / 97.8 \\
& Diffusion & 74.1 / 82.3 & 73.1 / 81.0 & 72.0 / 79.9 & 72.7 / 81.2 & 74.3 / 84.9 \\
\midrule
\multirow{2}{*}{CoDA} & GANs & \textbf{98.2} / \textbf{99.6} & \textbf{98.9} / \textbf{99.6} & \textbf{97.7} / 98.6 & \textbf{98.1} / \textbf{99.2} & \textbf{98.0} / \textbf{99.2} \\
& Diffusion & \textbf{95.3} / \textbf{98.7} & \textbf{95.1} / \textbf{98.4} & \textbf{94.8} / \textbf{98.2} & \textbf{93.7} / \textbf{98.1} & \textbf{94.4} / \textbf{98.0} \\
\bottomrule
\end{tabular}
\end{table*}

\begin{figure*}[!t]
\centering
\includegraphics[width=0.8\linewidth]{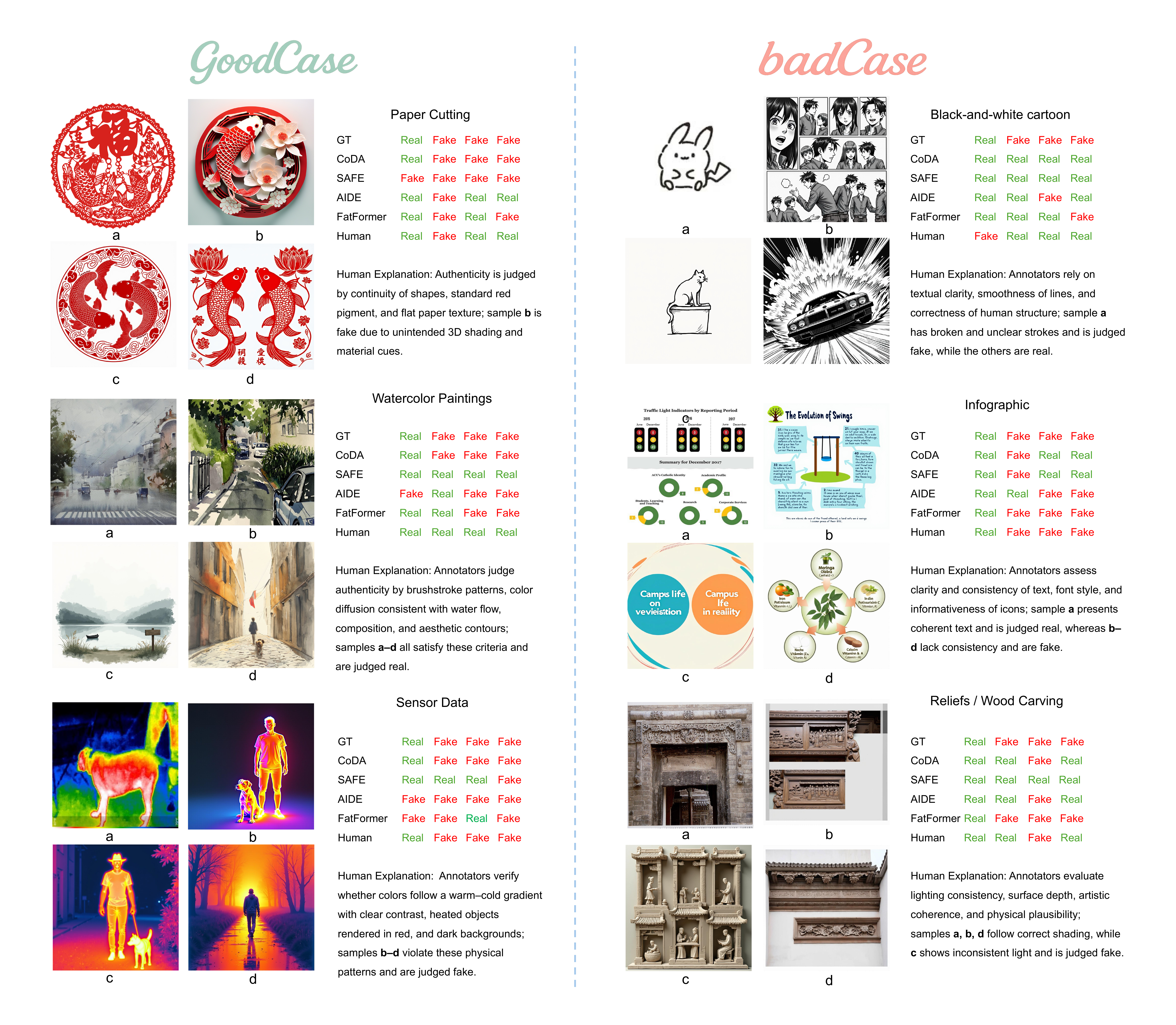}
\caption{Qualitative case study of CoDA. Examples in each group are arranged from (a) to (d), and GT denotes the ground-truth label. \textbf{Good cases (left):} the model performs well in color-rich domains by detecting subtle generative artifacts. \textbf{Failure cases (right):} limitations appear in domains with limited color cues or where higher-level reasoning about text, physical plausibility, or lighting consistency is required.}
\label{fig:case}
\end{figure*}

\begin{figure*}[t]
\centering
\includegraphics[width=0.8\textwidth]{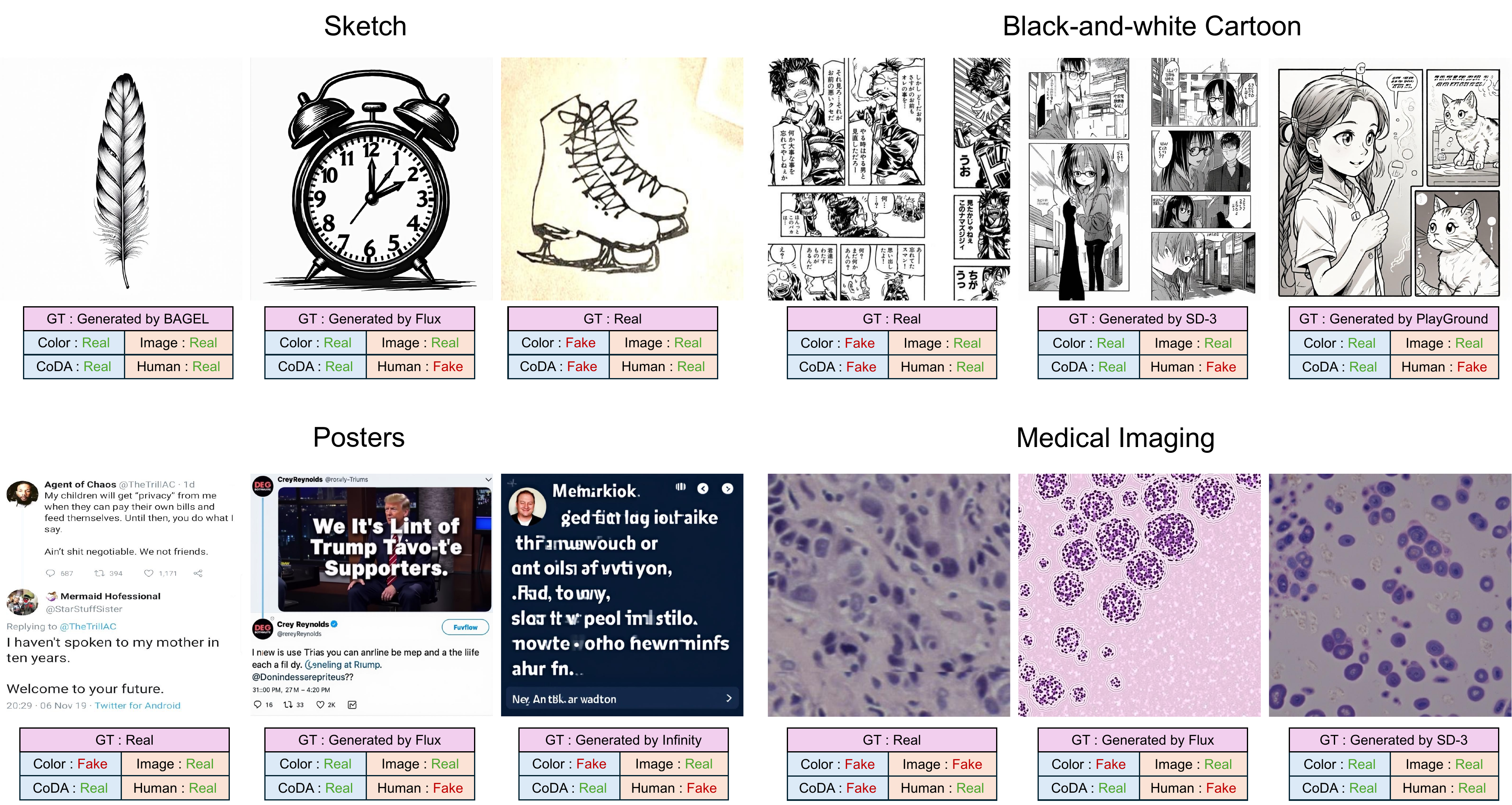}
\caption{Representative failure cases of CoDA across challenging FakeForm domains. For each case, we report the ground-truth label, the predictions of the color branch, the image branch, the full CoDA model, and the human judgment.}
\label{fig:badcase}
\end{figure*}

\subsection{Case Study and Failure Analysis}

Figure~\ref{fig:case} presents representative qualitative examples
illustrating both the strengths and limitations of CoDA across
diverse FakeForm domains. When generative artifacts manifest as
color-distribution irregularities, the Noise-Quantization Probe
yields clear and informative responses, enabling CoDA to make
accurate predictions while retaining the efficiency of a lightweight
detector.

Meanwhile, the failure cases in Figure~\ref{fig:badcase} reveal
scenarios where the color cue becomes less reliable.
In low-color cases such as \emph{Sketch} and black-and-white cartoons within the \emph{Cartoon} domain, chromatic evidence is inherently scarce, causing the color branch to lose discriminability and
shifting the burden to structural and stylistic cues captured by
the image branch.
In \emph{Poster}, the challenge moves to higher-level factors
such as text clarity, font consistency, and layout coherence.
In \emph{Medical Imaging}, domain-specific texture and appearance
statistics that deviate substantially from natural imagery further
amplify the difficulty.
These examples confirm that the effectiveness of the color cue is
domain-dependent, particularly under significant distribution shift.

More broadly, these cases delineate an important boundary of CoDA.
The method is designed to combine probe-induced color cues with
lightweight image features rather than to replace large semantic
models in tasks requiring fine-grained text understanding,
domain-specific expertise, or broader contextual reasoning.
Nonetheless, the qualitative results demonstrate that the
dual-branch design strikes a favorable balance between efficiency
and robustness: the image branch compensates when the color cue
weakens, while the color branch contributes useful evidence across
a wide range of domains.

\begin{table}[!t]
\centering
\caption{Ablation study results (Acc, \%).}
\label{tab:ablation}
\small
\setlength{\tabcolsep}{5.5pt}
\begin{tabular}{lcccc}
\toprule
Configuration & GANs & Ojha & GenImage & FakeForm \\
\midrule
\multicolumn{5}{l}{\textit{Component Analysis}} \\
Image Branch Only & 95.5 & 94.9 & 95.1 & 66.8 \\
Color Branch Only & 78.6 & 84.6 & 87.6 & 58.4 \\
Full Model & \textbf{98.7} & \textbf{97.5} & \textbf{95.9} & \textbf{77.7} \\
\midrule
\multicolumn{5}{l}{\textit{Noise Replicas ($R$)}} \\
$R = 20$ & 98.0 & 96.8 & 95.3 & 75.7 \\
$R = 30$ & 98.5 & 97.5 & 95.6 & 76.9 \\
$R = 50$ & \textbf{98.7} & \textbf{97.5} & \textbf{95.9} & \textbf{77.7} \\
\midrule
\multicolumn{5}{l}{\textit{Feature Extractors}} \\
CLIP Only & 96.1 & 95.3 & 94.7 & 75.2 \\
CLIP + Color & 98.7 & \textbf{97.8} & \textbf{96.1} & 76.9 \\
ResNet + Color & \textbf{98.7} & 97.5 & 95.9 & \textbf{77.7} \\
\midrule
\multicolumn{5}{l}{\textit{Noise Level ($\sigma$)}} \\
$\sigma = 0.05$ & 95.9 & 96.1 & 95.3 & 75.1 \\
$\sigma = 0.20$ & 96.4 & 96.9 & 95.6 & 75.3 \\
$\sigma = 0.10$ & \textbf{98.7} & \textbf{97.5} & \textbf{95.9} & \textbf{77.7} \\
\bottomrule
\end{tabular}
\end{table}

\begin{table}[t!]
\centering
\caption{Domain-wise branch contribution on FakeForm (Acc, \%). $\Delta$ denotes the gain over the Image Only branch.}
\label{tab:branch_domain}
\resizebox{\linewidth}{!}{
\begin{tabular}{l|ccc|c}
\toprule
\textbf{Domain Group} & \textbf{Color Only} & \textbf{Image Only} & \textbf{Full} & $\bm{\Delta}$ \\
\midrule
Photography          & 80.1 & 78.8 & 87.5 & +8.7 \\
Games                & 81.7 & 77.0 & 92.2 & +15.2 \\
Sketch               & 49.4 & 62.7 & 65.1 & +2.4 \\
Chinese Ink Painting & 58.6 & 67.9 & 70.2 & +2.3 \\
Cartoon (Color)      & 71.0 & 67.2 & 75.5 & +8.3 \\
Cartoon (B\&W)       & 54.5 & 66.9 & 68.2 & +1.3 \\
Traffic Maps         & 53.1 & 66.4 & 68.8 & +2.4 \\
\midrule
\textbf{Overall (all domains)} & 58.4 & 66.8 & 77.7 & +10.9 \\
\bottomrule
\end{tabular}
}
\end{table}

\begin{table}[t!]
\centering
\caption{Domain-wise sensitivity analysis on FakeForm (Acc, \%).}
\label{tab:hyper_domain}
\resizebox{0.98\linewidth}{!}{
\begin{tabular}{l|ccc|ccc}
\toprule
\multirow{2}{*}{\textbf{Domain Group}} 
& \multicolumn{3}{c|}{\textbf{Noise Replicas $R$} ($\sigma=0.10$)} 
& \multicolumn{3}{c}{\textbf{Noise Level $\sigma$} ($R=50$)} \\
& \textbf{$R=20$} & \textbf{$R=30$} & \textbf{$R=50$} 
& \textbf{$\sigma=0.05$} & \textbf{$\sigma=0.10$} & \textbf{$\sigma=0.20$} \\
\midrule
Photography          & 86.4 & 87.0 & 87.5 & 85.2 & 87.5 & 86.6 \\
Games                & 91.5 & 91.6 & 92.3 & 90.1 & 92.3 & 92.4 \\
Sketch               & 64.9 & 65.1 & 65.1 & 65.0 & 65.1 & 65.1 \\
Chinese Ink Painting & 69.5 & 70.0 & 70.2 & 69.6 & 70.2 & 70.2 \\
Cartoon (Color)      & 73.6 & 74.7 & 75.5 & 73.1 & 75.5 & 74.8 \\
Cartoon (B\&W)       & 68.0 & 68.2 & 68.2 & 67.9 & 68.2 & 68.3 \\
Traffic Maps         & 68.7 & 68.9 & 68.8 & 68.5 & 68.8 & 68.8 \\
\midrule
\textbf{Overall (all domains)} & 75.7 & 76.9 & 77.7 & 75.1 & 77.7 & 75.3 \\
\bottomrule
\end{tabular}
}
\end{table}

\subsection{Ablation Study}

To quantify the contribution of each component and justify the design choices, we conduct ablation studies on four benchmarks: ForenSynths (GANs), Ojha (Diffusion), GenImage, and FakeForm.

\textbf{Branch Complementarity and Feature Design.}
Table~\ref{tab:ablation} shows that the two branches play complementary roles. The Image Branch Only model performs strongly on standard benchmarks but drops to 66.8\% on FakeForm, while the Color Branch Only model is weaker overall and reaches 58.4\% in the cross-domain setting. In contrast, the Full Model consistently outperforms either single branch on all benchmarks and improves FakeForm accuracy from 66.8\% to 77.7\%. This result indicates that the color branch is not a standalone replacement for the image branch; rather, it provides useful complementary evidence that becomes especially valuable under domain shift.

We further compare the lightweight ResNet backbone with a pre-trained CLIP model. The CLIP Only model is already strong, achieving 75.2\% on FakeForm. Adding the color branch improves both backbones, and the proposed ResNet + Color combination achieves the best FakeForm result of 77.7\%. This suggests that features trained jointly with the probe-induced signal remain effective for cross-domain detection while preserving efficiency.

To better understand how the two branches behave across different forms of content, Table~\ref{tab:branch_domain} reports domain-wise branch contributions on representative FakeForm groups. The usefulness of the color branch is clearly domain-dependent. The gain of the full model over the image-only branch is largest in color-rich domains such as Games (+15.2), Photography (+8.7), and Cartoon (Color) (+8.3), while it becomes smaller in low-color domains such as Cartoon (B\&W) (+1.3), Sketch (+2.4), Chinese Ink Painting (+2.3), and Traffic Maps (+2.4). Nevertheless, the full model consistently outperforms the image-only branch in every listed group, confirming that the color branch provides complementary evidence even when it is not the dominant signal.

\textbf{Impact of Noise Replicas and Noise Level.}
The global ablation in Table~\ref{tab:ablation} shows that increasing the number of noise replicas from $R=20$ to $R=50$ steadily improves performance, with the strongest gain appearing on FakeForm. The noise level also has a clear influence: $\sigma=0.05$ under-perturbs the image, while $\sigma=0.20$ introduces excessive corruption. Across all four benchmarks, the default setting $(R=50,\sigma=0.10)$ yields the best overall trade-off.

To further examine whether these trends remain stable across domains, Table~\ref{tab:hyper_domain} reports a domain-wise sensitivity analysis on representative FakeForm groups. The overall trends remain highly consistent. For \textbf{Noise Replicas}, increasing $R$ generally improves performance, with $R=50$ achieving the best or tied-best result in all representative groups except Traffic Maps, where the difference is negligible (68.9 for $R=30$ versus 68.8 for $R=50$). For \textbf{Noise Level}, $\sigma=0.10$ is the most reliable overall choice. It achieves the best or tied-best result in Photography, Sketch, Chinese Ink Painting, and Cartoon (Color), and remains extremely close to the best setting in the remaining groups. Although Games and Cartoon (B\&W) show a marginal preference for $\sigma=0.20$, the gain is only 0.1 points in each case. More importantly, the overall performance across all domains is highest at $\sigma=0.10$ (77.7), compared with 75.1 for $\sigma=0.05$ and 75.3 for $\sigma=0.20$. These results indicate that the default setting $(R=50,\sigma=0.10)$ remains at or very close to the per-domain optimum across all representative domain groups.

\section{Conclusion and Future Work}

We presented CoDA, a lightweight detector for AI-generated images that combines a Noise-Quantization Probe with image features to exploit systematic differences in color distribution between real and generated content. We also introduced FakeForm, a large-scale benchmark of approximately 370,000 images across 62 diverse domains for evaluating both cross-model and cross-domain generalization. Experiments on existing benchmarks and FakeForm show that CoDA achieves strong accuracy, robustness to common perturbations, and a favorable balance between efficiency and generalization. These results indicate that color-distribution probing provides a useful complementary cue for robust AI-generated image detection.

Several directions remain for future work. First, the effectiveness of the color cue is domain-dependent and becomes weaker in low-color or highly specialized domains, suggesting the value of adaptive mechanisms or tighter integration with stronger semantic features. Second, extending the framework beyond static images to video and other emerging generative modalities may further broaden its applicability.

\bibliographystyle{IEEEtran}
\bibliography{mybib}

@inproceedings{karras2017progressive,
  title     = {Progressive Growing of {GAN}s for Improved Quality, Stability, and Variation},
  author    = {Karras, Tero and Aila, Timo and Laine, Samuli and Lehtinen, Jaakko},
  booktitle = {International Conference on Learning Representations (ICLR)},
  year      = {2018}
}

@inproceedings{karras2020analyzing,
  title     = {Analyzing and Improving the Image Quality of {StyleGAN}},
  author    = {Karras, Tero and Laine, Samuli and Aittala, Miika and Hellsten, Janne and Lehtinen, Jaakko and Aila, Timo},
  booktitle = {IEEE/CVF Conference on Computer Vision and Pattern Recognition (CVPR)},
  pages     = {8110--8119},
  year      = {2020}
}

@inproceedings{brock2018large,
  title     = {Large Scale {GAN} Training for High Fidelity Natural Image Synthesis},
  author    = {Brock, Andrew and Donahue, Jeff and Simonyan, Karen},
  booktitle = {International Conference on Learning Representations (ICLR)},
  year      = {2019}
}

@inproceedings{ho2020denoising,
  title     = {Denoising Diffusion Probabilistic Models},
  author    = {Ho, Jonathan and Jain, Ajay and Abbeel, Pieter},
  booktitle = {Advances in Neural Information Processing Systems (NeurIPS)},
  year      = {2020}
}

@inproceedings{podell2023sdxl,
  title     = {{SDXL}: Improving Latent Diffusion Models for High-Resolution Image Synthesis},
  author    = {Podell, Dustin and Sauer, Axel and Ke, Zheng and M{\"u}ller, Jonas and others},
  booktitle = {International Conference on Learning Representations (ICLR)},
  year      = {2024}
}

@inproceedings{wang2020cnn,
  title     = {{CNN}-Generated Images Are Surprisingly Easy to Spot... for Now},
  author    = {Wang, Sheng-Yu and Wang, Oliver and Zhang, Richard and Owens, Andrew and Efros, Alexei A.},
  booktitle = {IEEE/CVF Conference on Computer Vision and Pattern Recognition (CVPR)},
  pages     = {8695--8704},
  year      = {2020}
}

@inproceedings{frank2020leveraging,
  title     = {Leveraging Frequency Analysis for Deep Fake Image Recognition},
  author    = {Frank, Stephan and Andrus, Melanie and Jou, Brendan and Agrawala, Maneesh},
  booktitle = {International Conference on Machine Learning (ICML) Workshop},
  year      = {2020}
}

@inproceedings{tan2024frequency,
  title     = {Frequency-Aware Deepfake Detection: Improving Generalizability through Frequency Space Domain Learning},
  author    = {Tan, Chuangchuang and Zhao, Yao and Wei, Shikui and Gu, Guanghua and Liu, Ping and Wei, Yunchao},
  booktitle = {AAAI Conference on Artificial Intelligence (AAAI)},
  volume    = {38},
  number    = {5},
  pages     = {5052--5060},
  year      = {2024},
  doi       = {10.1609/aaai.v38i5.28310}
}

@article{zhong2023rich,
  title   = {{PatchCraft}: Exploring Texture Patch for Efficient {AI}-Generated Image Detection},
  author  = {Zhong, Nan and Xu, Yiran and Li, Sheng and Qian, Zhenxing and Zhang, Xinpeng},
  journal = {arXiv preprint arXiv:2311.12397},
  year    = {2023}
}

@inproceedings{ojha2023towards,
  title     = {Towards Universal Fake Image Detectors That Generalize Across Generative Models},
  author    = {Ojha, Utkarsh and Li, Yijun and Yang, Jing and Efros, Alexei A. and Zhang, Richard and Shechtman, Eli and Holynski, Aleksander},
  booktitle = {IEEE/CVF Conference on Computer Vision and Pattern Recognition (CVPR)},
  pages     = {24480--24490},
  year      = {2023}
}

@inproceedings{liu2024forgery,
  title     = {Forgery-Aware Adaptive Transformer for Generalizable Synthetic Image Detection},
  author    = {Liu, Zhen and Zhang, Kai and Wang, Wenqi and Zeng, Ailing and Tian, Yapeng and Luo, Ping and Liu, Ziwei},
  booktitle = {IEEE/CVF Conference on Computer Vision and Pattern Recognition (CVPR)},
  year      = {2024}
}

@article{finlayson2004intrinsic,
  title   = {Intrinsic Images by Entropy Minimization},
  author  = {Finlayson, Graham D. and Hordley, Steven D. and Drew, Mark S.},
  journal = {International Journal of Computer Vision},
  volume  = {56},
  number  = {3},
  pages   = {131--147},
  year    = {2004}
}

@article{vrhel1992color,
  title   = {Color Correction Using Principal Components},
  author  = {Vrhel, Michael J. and Trussell, H. Joel},
  journal = {{IEEE} Transactions on Image Processing},
  volume  = {1},
  number  = {4},
  pages   = {512--517},
  year    = {1992}
}

@book{foley1982fundamentals,
  title     = {Fundamentals of Interactive Computer Graphics},
  author    = {Foley, James D. and van Dam, Andries},
  publisher = {Addison-Wesley},
  year      = {1982}
}

@inproceedings{radford2021learning,
  title     = {Learning Transferable Visual Models From Natural Language Supervision},
  author    = {Radford, Alec and Kim, Jong Wook and Hallacy, Chris and Ramesh, Aditya and Goh, Gabriel and Agarwal, Sandhini and Sastry, Girish and Askell, Amanda and Mishkin, Pamela and Clark, Jack and Krueger, Gretchen and Sutskever, Ilya},
  booktitle = {Advances in Neural Information Processing Systems (NeurIPS)},
  year      = {2021}
}

@inproceedings{qian2020thinking,
  title     = {Thinking in Frequency: Face Forgery Detection by Mining Frequency-Aware Clues},
  author    = {Qian, Yu and Yin, Guo and Liu, Lu and Chen, Shouhong and Wang, Weiqiang},
  booktitle = {European Conference on Computer Vision (ECCV)},
  pages     = {86--103},
  year      = {2020}
}

@inproceedings{jeong2022bihpf,
  title     = {{BiHPF}: Bilateral High-Pass Filters for Robust Deepfake Detection},
  author    = {Jeong, Yujin and Choi, Hyun and Cho, Yunjey and Kim, Youngjung and Kim, Seunghoon and Yoon, Sungroh},
  booktitle = {European Conference on Computer Vision (ECCV)},
  pages     = {617--633},
  year      = {2022}
}

@inproceedings{ricker2022towards,
  title        = {Unveiling Universal Forensics of Diffusion Models with Adversarial Perturbations},
  author       = {Xie, Kangyang and Liu, Jiaan and Zhu, Muzhi and Ding, Ganggui and Liu, Zide and Chen, Hao and Chen, Hangyue},
  booktitle    = {International Joint Conference on Neural Networks (IJCNN)},
  pages        = {1--8},
  year         = {2024},
  organization = {{IEEE}}
}

@inproceedings{wang2023dire,
  title     = {{DIRE} for Diffusion-Generated Image Detection},
  author    = {Wang, Kai and Li, Xinyu and Wang, Xudong and Zhang, Richard and Zhu, Jun-Yan},
  booktitle = {IEEE/CVF International Conference on Computer Vision (ICCV)},
  year      = {2023}
}

@inproceedings{cui2024forensicsadapter,
  title     = {Forensics Adapter: Adapting {CLIP} for Generalizable Face Forgery Detection},
  author    = {Cui, Xinjie and Li, Yuezun and Luo, Ao and Zhou, Jiaran and Dong, Junyu},
  booktitle = {IEEE/CVF Conference on Computer Vision and Pattern Recognition (CVPR)},
  year      = {2025}
}

@article{qin2025infofd,
  title   = {Multimodal Conditional Information Bottleneck for Generalizable {AI}-Generated Image Detection},
  author  = {Qin, Haotian and Chang, Dongliang and Gao, Yueying and Yu, Bingyao and Chen, Lei and Ma, Zhanyu},
  journal = {arXiv preprint arXiv:2505.15217},
  year    = {2025}
}

@article{ho2022classifier,
  title   = {Classifier-Free Diffusion Guidance},
  author  = {Ho, Jonathan and Salimans, Tim},
  journal = {arXiv preprint arXiv:2207.12598},
  year    = {2022}
}

@inproceedings{saharia2022imagen,
  title     = {Photorealistic Text-to-Image Diffusion Models with Deep Language Understanding},
  author    = {Saharia, Chitwan and Chan, William and Saxena, Saurabh and Li, Lala and Whang, Jay and Denton, Emily L. and Ghasemipour, Seyed Kamyar Seyed and Ayan, Burcu Karagol and Mahdavi, S. Sara and Lopes, Raphael Gontijo and Salimans, Tim and Ho, Jonathan and Fleet, David J. and Norouzi, Mohammad},
  booktitle = {Advances in Neural Information Processing Systems (NeurIPS)},
  year      = {2022}
}

@inproceedings{lin2024wacv,
  title     = {Common Diffusion Noise Schedules and Sample Steps Are Flawed},
  author    = {Lin, Shanchuan and Liu, Bingchen and Li, Jiashi and Yang, Xiao},
  booktitle = {IEEE/CVF Winter Conference on Applications of Computer Vision (WACV)},
  pages     = {5404--5411},
  year      = {2024}
}

@inproceedings{sadat2024eliminating,
  title     = {Eliminating Oversaturation and Artifacts of High Guidance Scales in Diffusion Models},
  author    = {Sadat, Seyedmorteza and Hilliges, Otmar and Weber, Romann M.},
  booktitle = {International Conference on Learning Representations (ICLR)},
  year      = {2025}
}

@inproceedings{jin2025angle,
  title     = {Angle Domain Guidance: Latent Diffusion Requires Rotation Rather Than Extrapolation},
  author    = {Jin, Cheng and Xiao, Zhenyu and Liu, Chutao and Gu, Yuantao},
  booktitle = {International Conference on Machine Learning (ICML)},
  volume    = {267},
  series    = {Proceedings of Machine Learning Research},
  pages     = {28187--28212},
  year      = {2025}
}

@inproceedings{van2017neural,
  title     = {Neural Discrete Representation Learning},
  author    = {van den Oord, Aaron and Vinyals, Oriol and Kavukcuoglu, Koray},
  booktitle = {Advances in Neural Information Processing Systems (NeurIPS)},
  year      = {2017}
}

@inproceedings{ranzato2015sequence,
  title     = {Sequence Level Training with Recurrent Neural Networks},
  author    = {Ranzato, Marc'Aurelio and Chopra, Sumit and Auli, Michael and Zaremba, Wojciech},
  booktitle = {International Conference on Learning Representations (ICLR)},
  year      = {2016}
}

@inproceedings{zhu2024genimage,
  title     = {{GenImage}: A Million-Scale Benchmark for Detecting {AI}-Generated Images},
  author    = {Zhu, Zheng-Tian and Zhang, Hang and Ouyang, Junjie and Liu, Zifei and Chen, Zhanjie and Shen, Yuchao and Zhang, Shibao and Li, Jialu and Chen, Lumin and Wang, Chao and Zuo, Wenming and Liu, Ziwei},
  booktitle = {Advances in Neural Information Processing Systems (NeurIPS) Datasets and Benchmarks Track},
  year      = {2023}
}

@inproceedings{tan2023learning,
  title     = {Learning on Gradients: Generalized Artifacts Representation for {GAN}-Generated Images Detection},
  author    = {Tan, Chuangchuang and Zhao, Yao and Wei, Shikui and Gu, Guanghua and Wei, Yunchao},
  booktitle = {IEEE/CVF Conference on Computer Vision and Pattern Recognition (CVPR)},
  pages     = {12105--12114},
  year      = {2023}
}

@inproceedings{tan2024rethinking,
  title     = {Rethinking the Up-Sampling Operations in {CNN}-Based Generative Network for Generalizable Deepfake Detection},
  author    = {Tan, Chuangchuang and Zhao, Yao and Wei, Shikui and Gu, Guanghua and Liu, Ping and Wei, Yunchao},
  booktitle = {IEEE/CVF Conference on Computer Vision and Pattern Recognition (CVPR)},
  pages     = {28130--28139},
  year      = {2024}
}

@inproceedings{li2024improving,
  title     = {Improving Synthetic Image Detection towards Generalization: An Image Transformation Perspective},
  author    = {Li, Ouxiang and Cai, Jiayin and Hao, Yanbin and Jiang, Xiaolong and Hu, Yao and Feng, Fuli},
  booktitle = {ACM SIGKDD Conference on Knowledge Discovery and Data Mining (KDD)},
  year      = {2025},
  doi       = {10.1145/3690624.3709392}
}

@inproceedings{ramesh2021zero,
  title     = {Zero-Shot Text-to-Image Generation},
  author    = {Ramesh, Aditya and Pavlov, Mikhail and Goh, Gabriel and Gray, Scott and Voss, Chelsea and Radford, Alec and Chen, Mark and Sutskever, Ilya},
  booktitle = {International Conference on Machine Learning (ICML)},
  year      = {2021}
}

@article{yu2022scaling,
  title   = {Scaling Autoregressive Models for Content-Rich Text-to-Image Generation},
  author  = {Yu, Jiahui and Xu, Zihang and Koh, Jing Yu and Luong, Thang and Baid, Gaurav and Wang, Zirui and Vasudevan, Vijay and Ku, Alexander and Yang, Yinfei and Ayan, Burcu Karagol and Hutchinson, Ben and Han, Wei and Parekh, Zarana and Li, Xin and Zhang, Han and Baldridge, Jason and Wu, Yonghui},
  journal = {Transactions on Machine Learning Research (TMLR)},
  year    = {2022}
}

@inproceedings{nataraj2019detecting,
  title     = {Detecting {GAN}-Generated Fake Images Using Co-Occurrence Matrices},
  author    = {Nataraj, Lakshmanan and Mohammed, Tajuddin and Manjunath, B. S. and Chandrasekaran, Shivkumar and Flenner, Arjuna and Bappy, Jawadul H. and Roy-Chowdhury, Amit K.},
  booktitle = {Electronic Imaging},
  year      = {2019}
}

@inproceedings{yan2024sanity,
  title     = {A Sanity Check for {AI}-Generated Image Detection},
  author    = {Yan, Shilin and Li, Ouxiang and Cai, Jiayin and Hao, Yanbin and Jiang, Xiaolong and Hu, Yao and Xie, Weidi},
  booktitle = {International Conference on Learning Representations (ICLR)},
  year      = {2025}
}

@inproceedings{zhou2025aigi,
  title     = {{AIGI}-{H}olmes: Towards Explainable and Generalizable {AI}-Generated Image Detection via Multimodal Large Language Models},
  author    = {Zhou, Ziyin and Luo, Yunpeng and Wu, Yuanchen and Sun, Ke and Ji, Jiayi and Yan, Ke and Ding, Shouhong and Sun, Xiaoshuai and Wu, Yunsheng and Ji, Rongrong},
  booktitle = {IEEE/CVF International Conference on Computer Vision (ICCV)},
  year      = {2025}
}

@inproceedings{zhang2025d3qe,
  title     = {{D}$^3${QE}: Learning Discrete Distribution Discrepancy-Aware Quantization Error for Autoregressive-Generated Image Detection},
  author    = {Zhang, Yanran and Yu, Bingyao and Zheng, Yu and Zheng, Wenzhao and Duan, Yueqi and Chen, Lei and Zhou, Jie and Lu, Jiwen},
  booktitle = {IEEE/CVF International Conference on Computer Vision (ICCV)},
  year      = {2025}
}

@article{verdoliva2020media,
  title   = {Media Forensics and DeepFakes: An Overview},
  author  = {Verdoliva, Luisa},
  journal = {{IEEE} Transactions on Pattern Analysis and Machine Intelligence},
  volume  = {42},
  number  = {11},
  pages   = {910--932},
  year    = {2020}
}

@article{guarnera2020deepfake,
  title   = {DeepFake Detection by Analyzing Convolutional Traces},
  author  = {Guarnera, Luca and Giudice, Olivero and Battiato, Sebastiano},
  journal = {{IEEE} Transactions on Pattern Analysis and Machine Intelligence},
  volume  = {44},
  number  = {10},
  pages   = {6721--6734},
  year    = {2022}
}

@inproceedings{yu2022responsible,
  title     = {Responsible Disclosure of Generative Models Using Scalable Fingerprinting},
  author    = {Yu, Ning and Davis, Larry S. and Fritz, Mario},
  booktitle = {International Conference on Learning Representations (ICLR)},
  year      = {2022}
}

@inproceedings{yu2019attributing,
  title     = {Attributing Fake Images to {GANs}: Learning and Analyzing {GAN} Fingerprints},
  author    = {Yu, Ning and Davis, Larry S. and Fritz, Mario},
  booktitle = {IEEE/CVF International Conference on Computer Vision (ICCV)},
  pages     = {7556--7566},
  year      = {2019}
}

@inproceedings{rossler2019faceforensics++,
  title     = {FaceForensics++: Learning to Detect Manipulated Facial Images},
  author    = {R{\"o}ssler, Andreas and Cozzolino, Davide and Verdoliva, Luisa and Riess, Christian and Thies, Justus and Nie{\ss}ner, Matthias},
  booktitle = {IEEE/CVF International Conference on Computer Vision (ICCV)},
  pages     = {1--11},
  year      = {2019}
}

@inproceedings{lin2014coco,
  title     = {Microsoft {COCO}: Common Objects in Context},
  author    = {Lin, Tsung-Yi and Maire, Michael and Belongie, Serge and Bourdev, Lubomir and Girshick, Ross and Hays, James and Perona, Pietro and Ramanan, Deva and Zitnick, C. Lawrence and Doll{\'a}r, Piotr},
  booktitle = {European Conference on Computer Vision (ECCV)},
  pages     = {740--755},
  year      = {2014}
}

@article{Russakovsky2015ImageNet,
  title   = {{ImageNet} Large Scale Visual Recognition Challenge},
  author  = {Russakovsky, Olga and Deng, Jia and Su, Hao and Krause, Jonathan and Satheesh, Sanjeev and Ma, Sean and Huang, Zhiheng and Karpathy, Andrej and Khosla, Aditya and Bernstein, Michael and Berg, Alexander C. and Fei-Fei, Li},
  journal = {International Journal of Computer Vision},
  volume  = {115},
  number  = {3},
  pages   = {211--252},
  year    = {2015}
}

@article{yu2015lsun,
  title   = {{LSUN}: Construction of a Large-Scale Image Dataset Using Deep Learning with Humans in the Loop},
  author  = {Yu, Fisher and Zhang, Yinda and Song, Shuran and Seff, Ari and Xiao, Jianxiong},
  journal = {arXiv preprint arXiv:1506.03365},
  year    = {2015}
}

@article{wu2023human,
  title   = {Human Preference Score v2: A Solid Benchmark for Evaluating Human Preferences of Text-to-Image Synthesis},
  author  = {Wu, Xiaoshi and Hao, Yiming and Sun, Keqiang and Chen, Yixiong and Zhu, Feng and Zhao, Rui and Li, Hongsheng},
  journal = {arXiv preprint arXiv:2306.09341},
  year    = {2023}
}

@inproceedings{tang2025hart,
  title     = {{HART}: Efficient Visual Generation with Hybrid Autoregressive Transformer},
  author    = {Tang, Haotian and Wu, Yecheng and Yang, Shang and Xie, Enze and Chen, Junsong and Chen, Junyu and Zhang, Zhuoyang and Cai, Han and Lu, Yao and Han, Song},
  booktitle = {International Conference on Learning Representations (ICLR)},
  year      = {2025}
}

@inproceedings{zheng2025himar,
  title     = {Hierarchical Masked Autoregressive Models with Low-Resolution Token Pivots},
  author    = {Zheng, Guangting and Li, Yehao and Pan, Yingwei and Deng, Jiajun and Yao, Ting and Zhang, Yanyong and Mei, Tao},
  booktitle = {International Conference on Machine Learning (ICML)},
  volume    = {267},
  series    = {Proceedings of Machine Learning Research},
  pages     = {92928--92948},
  year      = {2025}
}

@inproceedings{tian2024var,
  title     = {Visual Autoregressive Modeling: Scalable Image Generation via Next-Scale Prediction},
  author    = {Tian, Keyu and Jiang, Yi and Yuan, Zehuan and Peng, Bingyue and Wang, Liwei},
  booktitle = {Advances in Neural Information Processing Systems (NeurIPS)},
  year      = {2024}
}

@inproceedings{han2025infinity,
  title     = {{Infinity}: Scaling Bitwise AutoRegressive Modeling for High-Resolution Image Synthesis},
  author    = {Han, Jian and Liu, Jinlai and Jiang, Yi and Yan, Bin and Zhang, Yuqi and Peng, Bingyue and Liu, Xiaobing},
  booktitle = {IEEE/CVF Conference on Computer Vision and Pattern Recognition (CVPR)},
  pages     = {29265--29274},
  year      = {2025}
}

@inproceedings{jia2025secret,
  title = {Secret Lies in Color: Enhancing {AI}-Generated Image Detection with Color Distribution Analysis},
  author    = {Jia, Zexi and Huang, Chuanwei and Zhu, Yeshuang and Fei, Hongyan and Duan, Xiaoyue and Yuan, Zhiqiang and Deng, Ying and Zhang, Jiapei and Zhang, Jinchao and Zhou, Jie},
  booktitle = {IEEE/CVF Conference on Computer Vision and Pattern Recognition (CVPR)},
  pages     = {13445--13454},
  year      = {2025}
}

@article{fang2026too,
  title={Too Vivid to Be Real? Benchmarking and Calibrating Generative Color Fidelity},
  author={Fang, Zhengyao and Jia, Zexi and Zhong, Yijia and Luo, Pengcheng and Zhang, Jinchao and Lu, Guangming and Yu, Jun and Pei, Wenjie},
  journal={arXiv preprint arXiv:2603.10990},
  year={2026}
}

@article{jia2026styledecoupler,
  title={StyleDecoupler: Generalizable Artistic Style Disentanglement},
  author={Jia, Zexi and Zhang, Jinchao and Zhou, Jie},
  journal={arXiv preprint arXiv:2601.17697},
  year={2026}
}

@inproceedings{jia2025visual,
  title={A visual leap in clip compositionality reasoning through generation of counterfactual sets},
  author={Jia, Zexi and Huang, Chuanwei and Fei, Hongyan and Zhu, Yeshuang and Yuan, Zhiqiang and Deng, Ying and Zhang, Jiapei and Zhang, Jinchao and Zhou, Jie},
  booktitle={Proceedings of the IEEE/CVF International Conference on Computer Vision},
  pages={23498--23507},
  year={2025}
}

@article{jia2026evaluating,
  title={Evaluating Generative Models via One-Dimensional Code Distributions},
  author={Jia, Zexi and Luo, Pengcheng and Zhong, Yijia and Zhang, Jinchao and Zhou, Jie},
  journal={arXiv preprint arXiv:2603.08064},
  year={2026}
}

\vspace{-10px}
\begin{IEEEbiography}[{\includegraphics[width=1in,height=1.25in,clip,keepaspectratio]{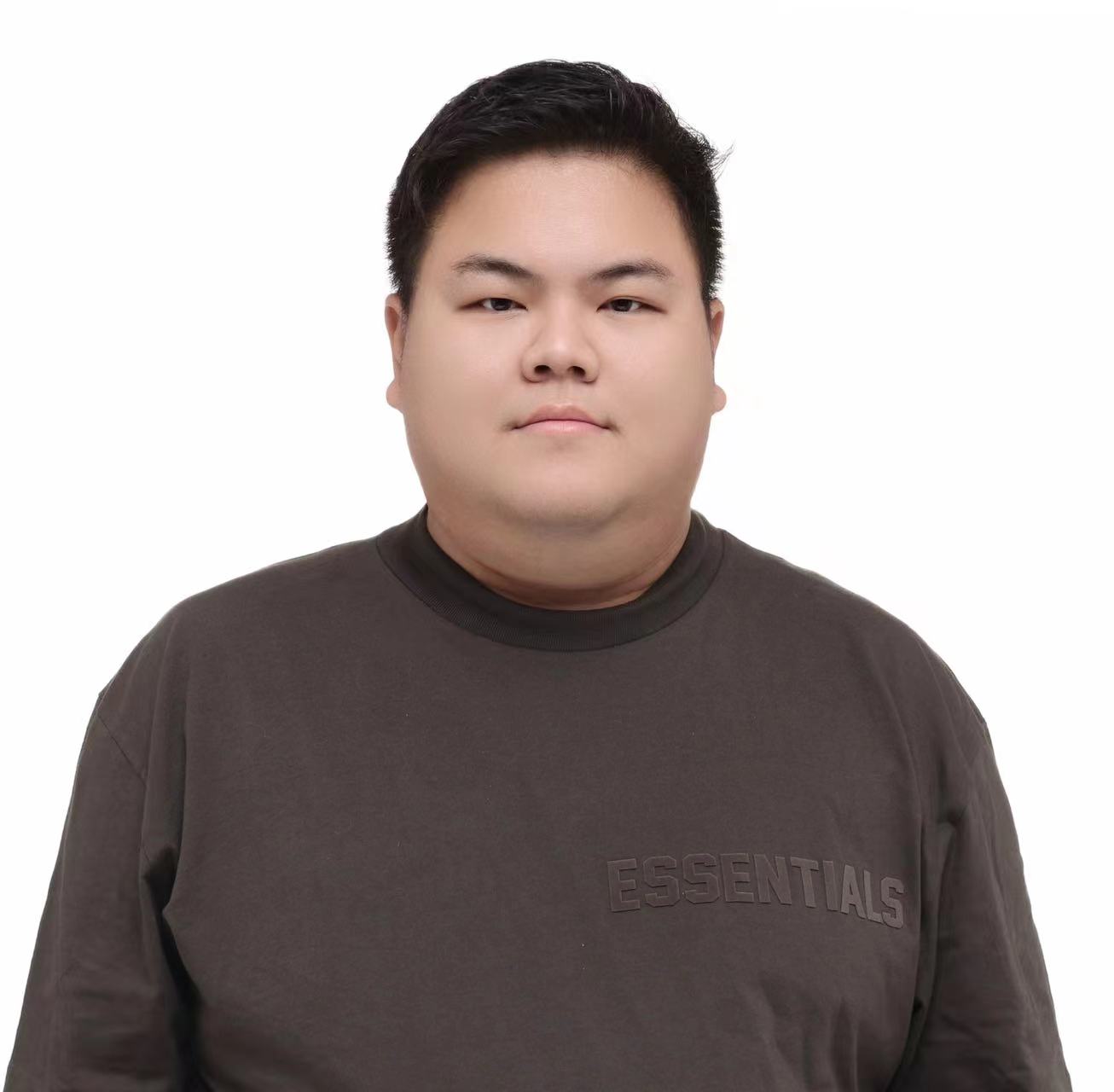}}]{Zexi Jia}
received the B.Eng. degree in software engineering (data mining) and the M.Sc. degree in computer science (computer vision) from Fudan University, Shanghai, China, in 2016 and 2020, respectively, and the Ph.D. degree in intelligent science (machine vision) from Peking University, Beijing, China, in 2024. He is currently a Senior Researcher with WeChat AI, Tencent, Beijing, China. His research interests include multimodal learning, vision-language models, AIGC safety, and biometric recognition.
\end{IEEEbiography}

\vspace{-10px}
\begin{IEEEbiography}[{\includegraphics[width=1in,height=1.25in,clip,keepaspectratio]{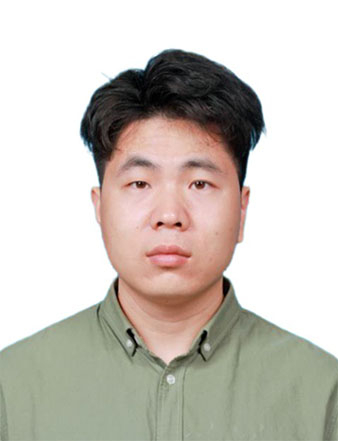}}]{Zhiqiang Yuan}
received the B.Sc. degree from Harbin Engineering University, Harbin, China, in 2019, and the Ph.D. degree from the Aerospace Information Research Institute, Chinese Academy of Sciences, Beijing, China, in 2024. He is currently a Senior Researcher with Tencent WeChat AI, Tencent, Beijing, China. His research interests include computer vision, pattern recognition, and generative models.
\end{IEEEbiography}

\begin{IEEEbiography}[{\includegraphics[width=1in,height=1.25in,clip,keepaspectratio]{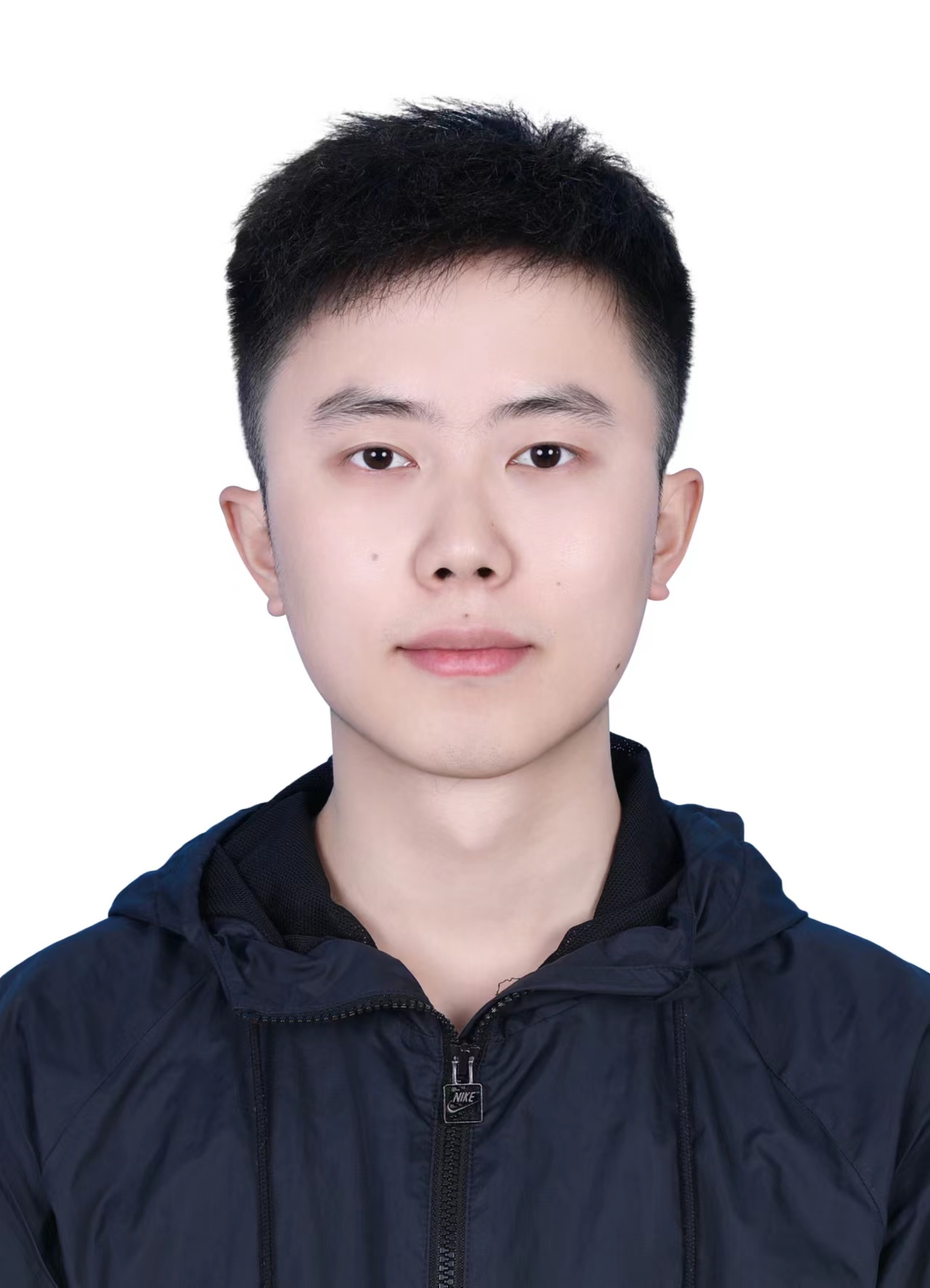}}]{Xiaoyue Duan}
received the bachelor's and master's degrees from the School of Automation Science and Electrical Engineering, Beihang University, Beijing, China, in 2021 and 2024, respectively. He is currently an Applied Researcher with Tencent WeChat AI, Beijing, China. His primary research interests include multimodal learning, vision-language models, image and video generation, and music generation.
\end{IEEEbiography}

\begin{IEEEbiography}[{\includegraphics[width=1in,height=1.25in,clip,keepaspectratio]{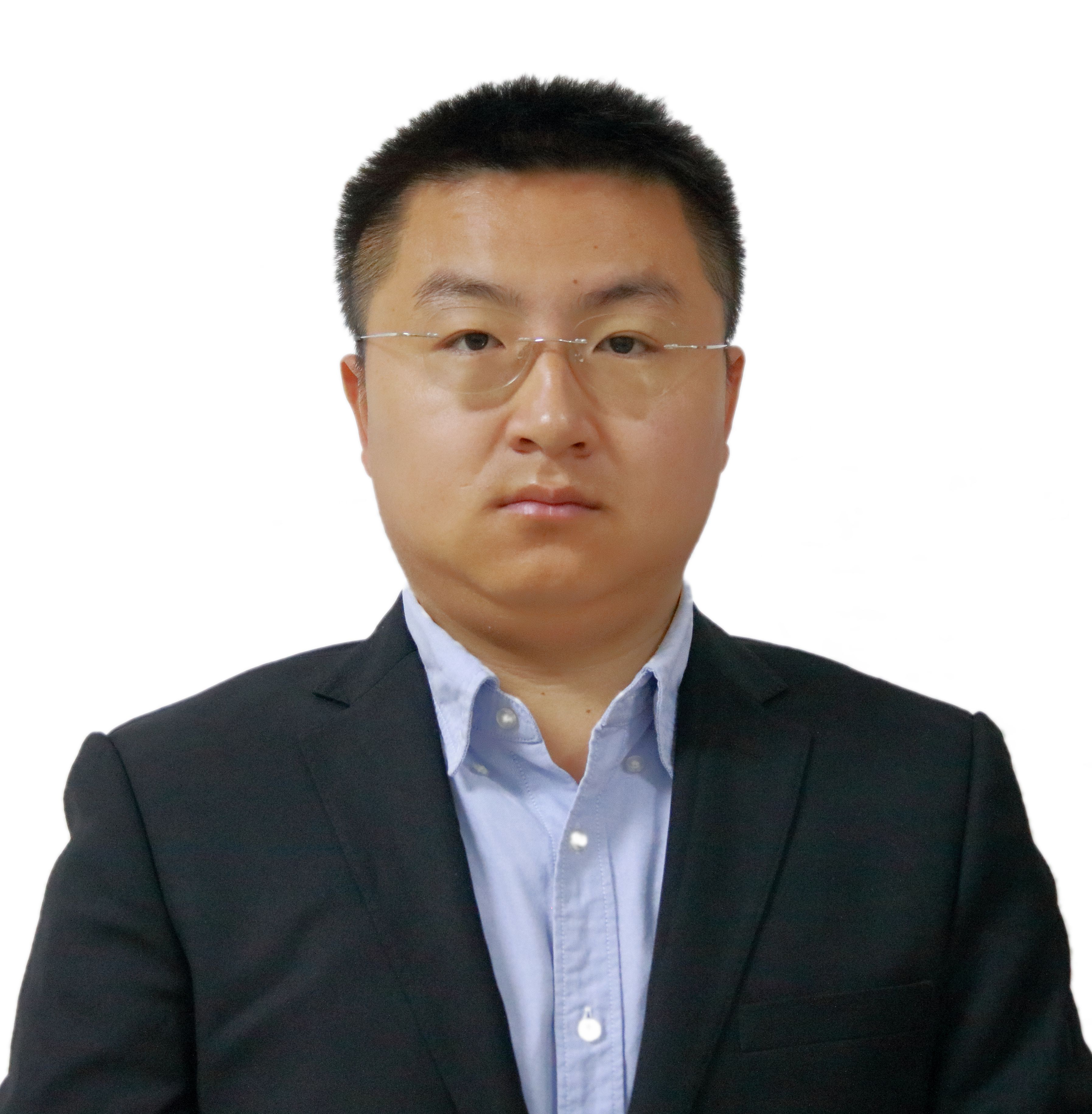}}]{Jinchao Zhang}
received the B.S. degree in computer science and technology from Linyi University, Linyi, China, in 2006, the M.S. degree in computer science and technology from Harbin Institute of Technology, Harbin, China, in 2013, and the Ph.D. degree in computer software from the University of Chinese Academy of Sciences, Beijing, China, in 2018. He is currently a Senior Researcher and Manager of the Multimodal Models Research Team with Tencent WeChat AI, Beijing, China. His research interests include machine learning and multimodal models. He has published over 70 peer-reviewed research papers at top-tier AI conferences and in reputable journals, including ACL, NeurIPS, CVPR, and AAAI.
\end{IEEEbiography}

\begin{IEEEbiography}[{\includegraphics[width=1in,height=1.25in,clip,keepaspectratio]{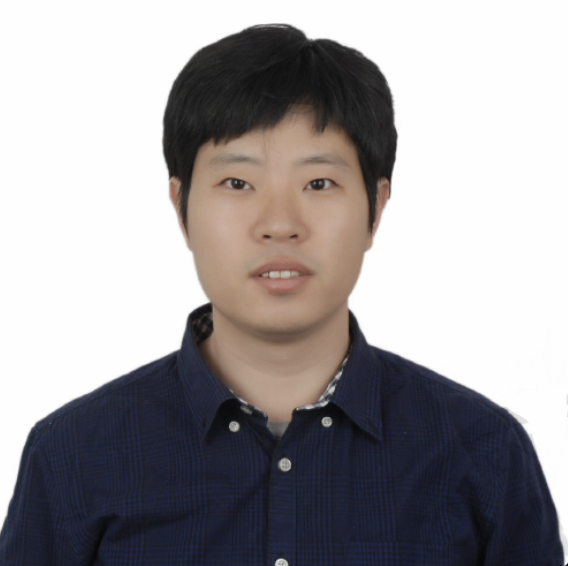}}]{Jie Zhou}
received the B.S. degree in physics from the University of Chinese Academy of Sciences, Hefei, China, in 2004, and the Ph.D. degree in statistical physics from the Institute of Theoretical Physics, Chinese Academy of Sciences, Beijing, China, in 2009. He is currently a Principal Researcher and Team Director with Tencent WeChat AI, Beijing, China. His research interests include natural language processing, deep learning, and their applications in conversational AI.
\end{IEEEbiography}

\begin{IEEEbiography}[{\includegraphics[width=1in,height=1.25in,clip,keepaspectratio]{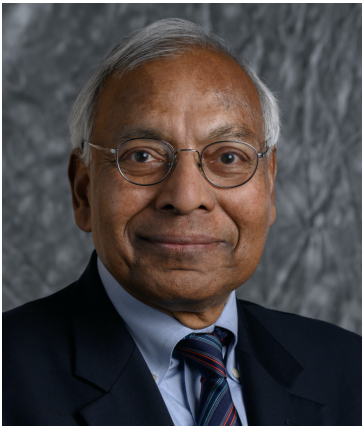}}]{Anil K. Jain}
is a University Distinguished Professor with the Department of Computer Science and Engineering, Michigan State University, East Lansing, MI, USA. His research interests include pattern recognition, computer vision, and biometric authentication. He served as the editor-in-chief of the IEEE Transactions on Pattern Analysis and Machine Intelligence and was a member of the United States Defense Science Board. He has received Fulbright, Guggenheim, Alexander von Humboldt, and IAPR King Sun Fu awards. He is a member of the National Academy of Engineering, the Indian National Academy of Engineering, the World Academy of Sciences, and the Chinese Academy of Sciences.
\end{IEEEbiography}

\vfill

\end{document}